\documentclass[final]{cvpr}

\usepackage{times}
\usepackage{epsfig}
\usepackage{graphicx}
\usepackage{amsmath}
\usepackage{amssymb}
\usepackage{multirow}
\usepackage{booktabs}
\usepackage[normalem]{ulem}
\usepackage[table,xcdraw,x11names]{xcolor}
\usepackage{placeins}
\usepackage{soul}                    % Color highlighting.
\usepackage{diagbox}  
\usepackage{lscape}
\usepackage{rotating}
\usepackage[pagebackref=true,breaklinks=true,letterpaper=true,colorlinks,bookmarks=false]{hyperref}
\expandafter\def\expandafter\UrlBreaks\expandafter{\UrlBreaks%  save the current one
  \do\a\do\b\do\c\do\d\do\e\do\f\do\g\do\h\do\i\do\j%
  \do\k\do\l\do\m\do\n\do\o\do\p\do\q\do\r\do\s\do\t%
  \do\u\do\v\do\w\do\x\do\y\do\z\do\A\do\B\do\C\do\D%
  \do\E\do\F\do\G\do\H\do\I\do\J\do\K\do\L\do\M\do\N%
  \do\O\do\P\do\Q\do\R\do\S\do\T\do\U\do\V\do\W\do\X%
  \do\Y\do\Z}

\DeclareMathOperator*{\argmin}{arg\,min}

\definecolor{darkred}{rgb}{0.6,0,0}
\definecolor{green}{rgb}{0.0,0.5,0}
\definecolor{blue}{rgb}{0,0,0.75}
\definecolor{orange}{rgb}{1,0.6,0.2}
\definecolor{red}{rgb}{1,0,0}
\definecolor{red2}{rgb}{1,0.2,0.1}
\definecolor{purplish}{rgb}{0.6,0,0.7}

\newcommand{\frev}[1]{\textcolor{red2}{#1}}

\def\real{\mathbb{R}}
\def\sdf{S} 
\def\brdf{E} 
\def\coords{\mathbf{x}}
\def\ro{\mathbf{r_o}}
\def\rd{\mathbf{r_d}}
\def\normal{\mathbf{n}}

\def\uv{\mathbf{u}}
\def\view{V}
\def\projection{P}
\def\images{\mathbf{I}}
\def\masks{\mathbf{M}}

\def\pixels{\mathbf{U}}
\def\siren{{\sc siren}}

\renewcommand{\frev}[1]{#1}

\graphicspath{
{figures/mini/}
{./data/images/}
} % where to search for the images

\def\cvprPaperID{4945} % *** Enter the CVPR Paper ID here
\def\confYear{CVPR 2021}

\newcommand{\beginsupplement}{%
        \setcounter{section}{0}
        \renewcommand{\thesection}{\Alph{section}}
     }
     
\newcommand{\beginacknowledgements}{%
        \renewcommand{\thesection}{}
     }

\begin{document}

%%%%%%%%% TITLE
\title{Neural Lumigraph Rendering}

\author{
Petr Kellnhofer$^{1,2}$,
Lars C. Jebe$^1$,
Andrew Jones$^1$,
Ryan Spicer$^1$,
Kari Pulli$^1$,
Gordon Wetzstein$^{2,1}$
\\
{$^1$Raxium $\quad\quad$ $^2$Stanford University} \\
{\tt\small \{pkellnhofer,ljebe,ajones,rspicer,kpulli,gwetzstein\}@raxium.com}
}

\maketitle
%\thispagestyle{empty}

%%%%%%%%% ABSTRACT
\begin{abstract}
Novel view synthesis is a challenging and ill-posed inverse rendering problem. Neural rendering techniques have recently achieved photorealistic image quality for this task. State-of-the-art (SOTA) neural volume rendering approaches, however, are slow to train and require minutes of inference (i.e., rendering) time for high image resolutions. We adopt high-capacity 
\frev{neural scene representations with periodic activations for jointly optimizing an implicit surface and a radiance field of a scene supervised exclusively with posed 2D images.} 
Our neural rendering pipeline accelerates SOTA neural volume rendering by about two orders of magnitude and our implicit surface representation is unique in allowing us to export a mesh with view-dependent texture information. Thus, like other implicit surface representations, ours is compatible with traditional graphics pipelines, enabling real-time rendering rates, while achieving unprecedented image quality compared to other surface methods. We assess the quality of our approach using existing datasets as well as high-quality 3D face data captured with a custom multi-camera rig.
\end{abstract}

%%%%%%%%% BODY TEXT

\section{Introduction}
\label{sec:introduction}
Novel view synthesis and 3D shape estimation from 2D images are inverse problems of fundamental importance in applications as diverse as photogrammetry, remote sensing, visualization, AR/VR, teleconferencing, visual effects, and games. While traditional 3D computer vision pipelines have been studied for decades, only emerging neural rendering techniques have been able to achieve photorealistic quality for novel view synthesis (e.g.,~\cite{mildenhall2020nerf,tewari2020state}). 

\begin{figure}[t!]
  \centering
  \includegraphics[width=\linewidth]{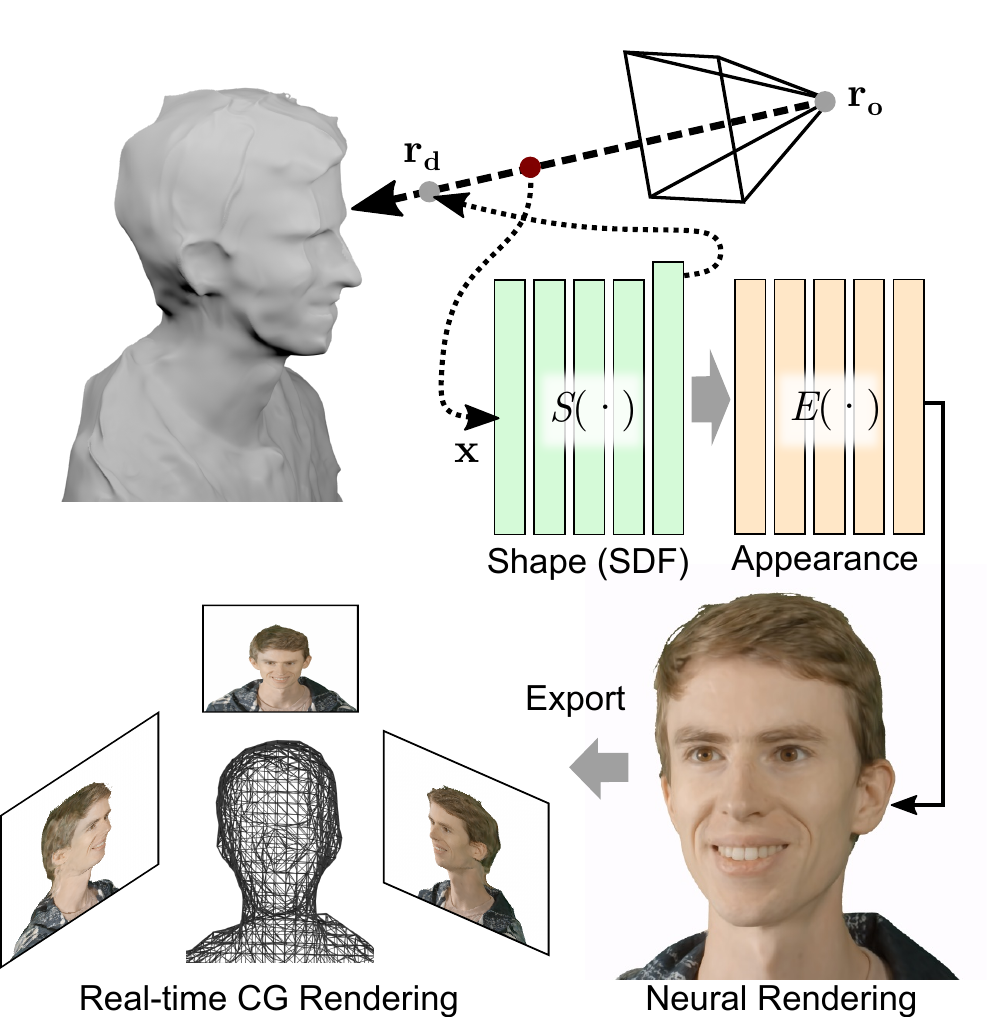}
  \caption{Overview of our framework. Given a set of multi-view images, we optimize representation networks modeling shape and appearance of a scene end to end using a differentiable sphere tracer. The resulting models can be exported to enable view-dependent real-time rendering using traditional graphics pipelines.	
	}
  \label{fig:model}
\end{figure}

State-of-the-art neural rendering approaches, such as neural radiance fields~\cite{mildenhall2020nerf}, however, do not offer real-time framerates, which severely limits their applicability to the aforementioned problems. This limitation is primarily imposed by the choice of implicit neural scene representation and rendering algorithm, namely a volumetric representation that requires a custom neural volume renderer. Neural surface representations, for example using signed distance functions (SDFs)~\cite{park2019deepsdf,gropp2020implicit,atzmon2019sal,yariv2020multiview}, occupancy fields~\cite{mescheder2019occupancy}, or feature-based representations~\cite{sitzmann2019srns}, on the other hand implicitly model the surface of objects, which can be extracted using the marching cubes algorithm~\cite{Lorensen:87} and exported into traditional mesh-based representations for real-time rendering. Although implicit neural surface representations have recently demonstrated impressive performance on shape reconstruction~\cite{yariv2020multiview}, their performance on view interpolation and synthesis tasks is limited. Thus, SOTA neural rendering approaches either perform well for view synthesis~\cite{mildenhall2020nerf} or 3D shape estimation~\cite{yariv2020multiview}, but not both.

Here, we adopt an SDF-based sinusoidal representation network ({\sc siren}) as the backbone of our neural rendering system. While these representations have recently demonstrated impressive performance on representing shapes via direct 3D supervision with point clouds~\cite{sitzmann2020siren}, we are the first to demonstrate how to leverage {\sc siren}'s extreme capacity in the context of learning 3D shapes using 2D supervision with images via neural rendering. For this purpose, we devise a novel loss function that maintains {\sc siren}'s high-capacity encoding for the supervised images while constraining it in the angular domain to prevent overfitting on these views. This training procedure allows us to robustly fit a {\sc siren}-based SDF directly to a sparse set of multi-view images. Our 2D-supervised implicit neural scene representation and rendering approach performs on par with NeRF on view interpolation tasks while providing a high-quality 3D surface that can be directly exported for real-time rendering at test time.  

Specifically, we make the following contributions: 
\begin{itemize}
	\item We develop a neural rendering framework comprising an implicit neural 3D scene representation, a neural renderer, and a custom loss function for training. This approach achieves $10 \times$ higher rendering rates than NeRF while providing comparable, SOTA image quality with the additional benefit of optimizing an implicitly defined surface. 
	\item We demonstrate how both shape and view-dependent appearance of our neural scene representation can be exported and rendered in real time using traditional graphics pipelines.
	\item We also build a custom camera array and capture several datasets of faces and heads for evaluating our approach and baselines. These data \frev{are available on the project website.}\footnote{\protect\url{http://www.computationalimaging.org/publications/nlr/}}  
\end{itemize}

\section{Related Work}
\label{sec:related}

% 1 paragraph on traditional 3D computer vision pipelines, cite Szeliski and COLMAP
Traditional 3D computer vision pipelines use structure-from-motion and multi-view-stereo algorithms to estimate sparse point clouds, camera poses, and textured meshes from 2D input views (e.g.,~\cite{Szeliski:book,Seitz:2006,Collet:2015,schonberger2016structure,schoenberger2016mvs}). Re-rendering these scene representations, however, does not achieve photorealistic image quality. As an alternative, image-based rendering techniques have been explored for decades~\cite{shum2000review}. 
\frev{Lumigraph rendering~\cite{Gortler:1996,buehler2001unstructured} stands out among these methods as an approach that leverages proxy scene geometry to interpolate the captured views better.}   
Still, these traditional approaches have not demonstrated photorealistic view synthesis for general 3D scenes. 

Emerging neural scene representations often model an object or scene explicitly using some 3D proxy geometry, such as an imperfect mesh~\cite{hedman2018deep,thies2019deferred,riegler2020free,zhang2020neural} or depth map~\cite{penner2017soft3d} estimated by multi-view stereo or other means, an object-specific shape template~\cite{kanazawa2018learning}, a multi-plane~\cite{Zhou:2018,Mildenhall:2019,flynn2019deepview} or multi-sphere~\cite{Broxton:2020,Attal:2020:ECCV} image, or a volume~\cite{sitzmann2019deepvoxels,Lombardi:2019}. An overview of recent neural rendering techniques, including extensive discussions of explicit representations, is provided by Tewari et al.~\cite{tewari2020state}. 

As opposed to explicit representations, emerging neural implicit scene representations promise 3D-structure-aware, continuous, memory-efficient representations for shape parts~\cite{genova2019learning,genova2019deep}, objects~\cite{park2019deepsdf,michalkiewicz2019implicit,atzmon2019sal,gropp2020implicit,yariv2020multiview,davies2020overfit,chabra2020deep}, or scenes~\cite{eslami2018neural,sitzmann2019srns,jiang2020local,peng2020convolutional,sitzmann2020siren}. These representations implicitly define an object or a scene using a neural network and can be supervised directly with 3D data, such as point clouds, or with 2D multi-view images~\cite{saito2019pifu,sitzmann2019srns,Oechsle2019ICCV,Niemeyer2020CVPR,mildenhall2020nerf,yariv2020multiview,liu2020neural,jiang2020sdfdiff,liu2020dist,kohli2020inferring}. It is important to distinguish between neural implicit representations that use implicitly defined volumes~\cite{mildenhall2020nerf,liu2020neural,lindell2020autoint} from those using implicitly defined surfaces, for example represented as signed distance functions (SDFs)~\cite{park2019deepsdf,michalkiewicz2019implicit,atzmon2019sal,gropp2020implicit,sitzmann2019srns,jiang2020local,peng2020convolutional} or occupancy networks~\cite{mescheder2019occupancy,chen2019learning,Niemeyer2020CVPR}. Surface-based representations allow for traditional mesh representations to be extracted and rendered efficiently with traditional computer graphics pipelines.

The neural rendering methods closest to our work are neural radiance fields (NeRF)~\cite{mildenhall2020nerf}, which provide the best image quality for view synthesis to date but do not directly model object shape, and implicit differentiable renderer (IDR)~\cite{yariv2020multiview}, which recently demonstrated SOTA performance for shape estimation but which does not achieve the same quality as NeRF for view synthesis. Our work leverages emerging sinusoidal representation networks ({\sc siren})~\cite{sitzmann2020siren} to achieve both of these capabilities simultaneously. This is crucial for the neural rendering pipeline we propose, which first learns a neural implicit surface representation and then exports it into a format that is compatible with existing real-time graphics pipelines. Although {\sc siren}s have been proposed in prior work~\cite{sitzmann2020siren}, we are the first to demonstrate how to leverage their impressive capacity with 2D multi-view image supervision -- a non-trivial task due to {\sc siren}'s extreme overfitting behavior.

\section{Neural Rendering Pipeline}
\label{sec:neuralrendering}
In this section, we describe our differentiable neural rendering pipeline, which is illustrated in Fig.~\ref{fig:model}.

%%%%%%%%%%%%%%%%%%%%%%%%%%%%%%%%%%%%%%%%%%%%%%%%%%%%%%%%%%%%%%%%%%%%%%%%%%%%%%%%%%%%%%%%%%%%
\subsection{Representation} 

We represent both shape and appearance of 3D objects using implicit functions in a framework similar to IDR~\cite{yariv2020multiview}. Unlike previous work, however, our network architecture builds on sinusoidal representation networks (\siren)~\cite{sitzmann2020siren}, which allow us to represent signals of significantly higher complexity within the same number of learnable parameters compared with common non-periodic multilayer perceptrons (MLP). 

We express the continuous shapes of a scene as the zero-level set $\sdf_0 = \{ \coords \, | \, \sdf(\coords) = 0\}$ of a signed distance function (SDF) 
\begin{align}
\sdf(\coords; \theta) : \real^3 \to \real,
\end{align}
where $\coords \in \real^3$ is a location in 3D space and $\theta$ are the learnable parameters of our \siren-based SDF representation.

Next, we model appearance as a spatially varying emission function, \frev{or radiance field}, $\brdf$ for directions $\rd \in \real^3$ defined in a global coordinate system. This formulation does not allow for relighting but it enables photorealistic reconstruction of the appearance of a scene under fixed lighting conditions. We leave the problem of modeling lighting and shading as an avenue of future work.

We additionally condition  $\brdf$ by the local normal direction $\normal = \nabla_\coords \sdf(\coords)$ as computed by automatic differentiation. This does not constrain any degrees of freedom but it has been shown to improve the training performance \cite{yariv2020multiview}. 
Finally, we also reuse $\theta$ to increase the network capacity and allow for modeling of fine spatial details and micro-reflections that are of a notably higher spatial complexity than the underlying shape. 
Together, we express the \frev{radiance field} as 
\begin{align}
\brdf(\coords, \rd, \normal; \theta, \phi) : \real^9 \to \real^3,
\end{align}
to represent RGB appearance using the additional learnable parameters $\phi$.

%%%%%%%%%%%%%%%%%%%%%%%%%%%%%%%%%%%%%%%%%%%%%%%%%%%%%%%%%%%%%%%%%%%%%%%%%%%%%%%%%%%%%%%%%%%%
\subsection{Neural Rendering} 

The goal of neural rendering is to project a 3D neural scene representation into one or multiple 2D images. We solve this task in two steps: 1) We find the 3D surface as the zero-level set $\sdf_0$ closest to the camera origin along each ray; 2) We resolve the appearance by sampling the local \frev{radiance} $\brdf$.

To address 1), we sphere trace the SDF to find $\sdf_0$~\cite{hart1996sphere}. 
For this purpose, we define a view and a projection matrix, $\view \in \real^{4 \times 4}$ and $\projection \in \real^{4 \times 4}$, similar to OpenGL's rendering API~\cite{woo1999opengl}. 
A ray origin $\ro$ and direction $\rd$ for an output pixel at relative projection plane location $\uv \in [-1,1]^2$ is then
\begin{align}
\ro & = (\view^{-1} \cdot [0,0,1,0]^T)_{x,y,z}, \\
\rd & = \nu\left((\projection \cdot \view)^{-1} \cdot [\uv_x, \uv_y, 0, 1]^T\right),
\end{align}
where $(\cdot)_{x,y,z} $ are vector components and $\nu(\mathbf{\omega}) = \mathbf{\omega}_{x,y,z} / ||\mathbf{\omega}_{x,y,z}||$ is vector normalization.

The sphere-tracing algorithm minimizes $|\sdf(\coords,\theta)|$ along each ray using iterative updates of the form
\begin{align}
\coords_0 = \ro, \quad \coords_{i+1} = \coords_{i} + \sdf(\coords_i) \rd.
\end{align}
Finally, $\sdf_0 = \{\coords_n\,|\,\sdf(\coords_n) = 0\}$ is the zero-set of rays converged to a foreground object for the step count $n = 16$. 
A small residual $|\sdf(\coords_n)|<0.005$ is tolerated in practice.
As proposed in recent work~\cite{yariv2020multiview,jiang2020sdfdiff,liu2020dist}, we only retain gradients in the last step rather than for all steps of the sphere tracer. This approach makes sphere tracing memory efficient.
Please refer to the supplemental materials for additional details.

The appearance is directly sampled from our \frev{radiance field} as $\brdf(\sdf_0, \rd, \nabla\sdf(\sdf_0); \theta, \phi)$.

%%%%%%%%%%%%%%%%%%%%%%%%%%%%%%%%%%%%%%%%%%%%%%%%%%%%%%%%%%%%%%%%%%%%%%%%%%%%%%%%%%%%%%%%%%%%
\subsection{Loss Function}

We supervise our 3D representation using a set of $m$ multi-view 2D images $\images = \real^{m \times w \times h \times 3}$ with known object masks $\masks = \real^{m \times w \times h}$ where 1 marks foreground. Our unique approach to leveraging \siren{} as a neural representation in this setting is challenging, because of \siren's tendency to overfit the signal to the supervised views.

In total, we use four different constraints to optimize the end-to-end representation using mini-batches of image pixels $\pixels$ with RGB values $\images_\pixels$ and object masks $\masks_\pixels$.

First, we minimize an $L1$ image reconstruction error for the true foreground pixels $\pixels_f = \pixels \cap \sdf_0 \cap \{\pixels\,|\,\masks_\pixels = 1\}$ as
\begin{align}
\mathcal{L}_R = \frac{1}{|\pixels|} \sum_{\mathbf{c} \in \images_{\pixels_f}} {|\brdf(\coords, \rd, \normal; \theta, \phi) - \mathbf{c}|},
\end{align}
where $\mathbf{c}$ is an RGB value of a foreground pixel in a minibatch. Both $L1$ and $L2$ work well but we have found $L1$ to produce marginally sharper images.

Second, we regularize the $\sdf$ by an eikonal constraint
\begin{align}
\mathcal{L}_E = \frac{1}{|\pixels|} \sum_{\coords_r} \big\|(\|\nabla_\coords \sdf(\coords_r;\theta)\|_2 - 1)\big\|_2^2
\end{align}
to enforce its metric properties important for efficient sphere tracing~\cite{icml2020_2086,jiang2020sdfdiff,sitzmann2020siren,yariv2020multiview}. Random points $\coords_r$ are uniformly sampled from a cube which encapsulates the object's bounding unit radius sphere.

Third, we restrict the coarse shape by enforcing its projected pattern to fall within the boundaries of the object masks. For this purpose, we adopt the soft mask loss proposed in~\cite{yariv2020multiview} defined for the non-foreground pixels and softness parameter $\alpha$ as
\begin{align}
\mathcal{L}_M = \frac{1}{\alpha |\pixels|} \sum_{m \in \masks_{\pixels \setminus \pixels_f}} \textrm{\sc bce}(\textrm{sigmoid}(-\alpha \sdf_{min}), m),
\end{align}
where $\textrm{\sc bce}$ is the binary cross entropy and $\sdf_{min} = \argmin_t{\sdf(\ro+t\rd;\theta)}$ is the minimum $\sdf$ value along the entire ray approximated by dense sampling of $t$.

Finally, we regularize the \frev{radiance field} $\brdf$ to avoid overfitting to training views. 
{\siren}s have a remarkable regressive potential, which biases them to overfit the appearance to the training views.
We leverage this power to allow for encoding of photorealistic surface details, but we need to restrict the behavior of the $\brdf$ in the angular domain conditioned by $\rd$ to achieve favorable interpolation behavior.
Inspired by multi-view projective texture mapping~\cite{debevec1998efficient}, we linearize the angular behavior using a smoothness term
\begin{align}
\mathcal{L}_S = \frac{1}{|\pixels|} \sum ||\nabla_{\rd}^2 \brdf(\coords, \rd, \normal; \theta, \phi)||_2^2.
\end{align}
Note that such level of control is unique to \siren{} and related architectures as they are $C^{\infty}$ differentiable.

Together, we optimize parameters $\theta$ and $\phi$ as
\begin{align}
\argmin_{\theta,\phi} \mathcal{L}_R + w_E \mathcal{L}_E + w_M \mathcal{L}_M + w_S \mathcal{L}_S,
\end{align}
with weights $w_E = 0.1$, $w_M = 100$, and $w_S = 0.01$ for all of our experiments. We have not found the performance to be very sensitive to this choice with the exception of $w_S$ where large values cause high-frequency artifacts in $\sdf$.

%%%%%%%%%%%%%%%%%%%%%%%%%%%%%%%%%%%%%%%%%%%%%%%%%%%%%%%%%%%%%%%%%%%%%%%%%%%%%%%%%%%%%%%%%%%%
\subsection{Additional Training Details}

We optimize the loss in mini-batches of 50,000 individual rays sampled uniformly across the entire training dataset. We have found a large batch size and uniform ray distribution to be critical to prevent local overfitting of \siren, especially for the high-frequency function $\brdf$.

We implement the MLPs representing $\sdf$ and $\brdf$ as {\siren}s with 5~layers using 256 hidden units each.
Additionally, we use Fourier features $\{\sin(2k\pi\rd),\cos(2k\pi\rd)\,|\,k\in1\ldots4\}$ in $\brdf$ to further support angular resolution \cite{mildenhall2020nerf,yariv2020multiview}.
\frev{This strategy is necessary to fit the sparsely supervised rays well while $\mathcal{L}_S$ enhances interpolation between them.}

We initialize $\sdf$ to a unit sphere of radius $0.5$ by pre-training to a procedural shape as described in~\cite{sitzmann2020siren}. 
We trace the object rays in a larger sphere of radius $1$,
but we have found that the smaller initial radius improves the initial fit as well as the consequent convergence rate.

We implement our method in PyTorch~\cite{NEURIPS2019_9015} and optimize the loss using the Adam solver~\cite{kingma2014adam} with an initial learning rate of $10^{-4}$ decreased by a factor of $2$ every 40,000 batches for the overall training length of 150,000 batches on a single Nvidia GPU RTX 2080Ti.

\section{Real-time Rendering Pipeline}
\label{sec:realtimerendering}
While we show that \siren{} is remarkably efficient in shape and appearance representation with 2D supervision, the required sphere tracer does not run at real-time rates for moderate to high image resolutions.
To overcome this challenge, we show how to leverage the compactness of our surface-based representation and convert our neural model to a triangular mesh suitable for real-time computer graphics applications.
For this purpose, we leverage unstructured lumigraph rendering, which preserves view-dependent effects learned by our neural representation~\cite{buehler2001unstructured}.

\subsection{Mesh extraction}

First, we use the marching cubes algorithm \cite{lorensen1987marching} to extract a high-resolution surface mesh from the SDF $\sdf$ voxelized at a resolution of $512^3$. Instead of extracting the zero-level set, we found that offsetting the iso-surface of $\sdf$ by \frev{$0.5\%$ of the object radius in the outside direction} optimizes the resulting image quality in practice.
To export the appearance, we resample the optimized emissivity function $\brdf$ to synthesize projective textures $\mathbf{T}_i$ for $N$ camera poses and corresponding projection matrices. 
\frev{The ability to resample the camera poses for efficient viewing space coverage is a key feature of our method and we explore the choice of $N$ and camera distributions in the supplement.}

\subsection{Rendering}

First, we rasterize the extracted mesh using OpenGL~\cite{woo1999opengl} and project the vertex positions to each pixel.
Next, we compute angles $\tau_{1 \dots N}$ between the ray towards the \frev{current} rendering camera and the rays towards each of the $N$ projective texture map viewpoints.
We then apply the unstructured lumigraph rendering technique of Buehler~et~al.~\cite{buehler2001unstructured} to blend contributions from the first $k=5$ textures, sorted by $\tau_i$ in ascending order, yielding a rendered image
\begin{align}
\mathbf{R} = \textstyle \sum_{i=1\dots k} w_i \mathbf{T}_i,
\end{align}
where the weights $w_i$ are computed as
\begin{align}
\hat{w}_i &= 1/\tau_i (1 - \tau_i/\tau_k), \\
w_i &= \hat{w}_i/\textstyle \sum_{i=1\dots k} \hat{w}_i.
\end{align}
This formulation satisfies the epipolar consistency by converging to an exclusive mapping by texture $\mathbf{T}_j$ when $\tau_j~\to~0$ \cite{buehler2001unstructured}.
Additionally, we discard samples from occluded textures by setting their $w_i$ to zero.
Occlusions are detected by a comparison between the pre-rendered depth associated with a texture and the distance between the mesh voxel and the texture viewpoint.
The same technique is commonly used in real-time graphics for shadow mapping.

\subsection{Evaluation}
\label{sec:ras_eval}

\begin{table}[]
\footnotesize
\centering
\useunder{\uline}{\ul}{}
\begin{tabular}{lrr}
\hline
Method                            & Render time [s]    & Model size [MB] \\ \hline
Colmap~\cite{schoenberger2016mvs} & \textbf{Real-time} & 30.39            \\
IDR~\cite{yariv2020multiview}     & 45                 & 11.13            \\
NV~\cite{Lombardi:2019}           & 0.65               & 438.36           \\
NeRF~\cite{Mildenhall:2019}       & 150                & 2.27             \\
NLR-ST                         & 13                 & \textbf{2.07}    \\
NLR-RAS                         & \textbf{Real-time} & 34.68            \\ \hline
\end{tabular}
\caption{Rendering time and representation size comparison for the DTU scan 65 \cite{jensen2014large} at 1600 $\times$ 1200 pixel resolution. 
\frev{``Real-time'' denotes framerates of at least 60~fps.}
}
  \label{tbl:performance}
\end{table}

\paragraph{Efficiency}
We compare the efficiency of our real-time \frev{rasterized neural lumigraph renderer (NLR-RAS)} with our \frev{sphere-traced renderer (NLR-ST}, Sec.~\ref{sec:neuralrendering}) along with other baselines in Table~\ref{tbl:performance}.
We observe, that although both \frev{NLR-ST} and IDR are based on \frev{sphere tracing}, the capacity of \siren{} allows for a smaller and faster model, which is evident by the model size.
Furthermore, the results show how costly the implicit volumetric rendering is.
In conclusion, only the explicit representations of Colmap and our \frev{NLR-RAS} allow for truly real-time performance with framerates over 60~fps at HD resolution on commodity hardware.

\paragraph{Image quality}
Both the quantitative comparisons in Tables~\ref{tbl:dtu}, \ref{tbl:others} and qualitative examples in Figures~\ref{fig:results}, \ref{fig:views} demonstrate the high \frev{NLR-RAS} rendering quality.
While lower than that of our \frev{NLR-ST} renderer, the \frev{NLR-RAS} still achieves PSNRs far superior to other explicit (Colmap) and implicit (IDR) surface representations.

\section{Camera Array and Data}
\label{sec:cameraarray}
\begin{figure}
	\centering
	\includegraphics[width=0.7\linewidth]{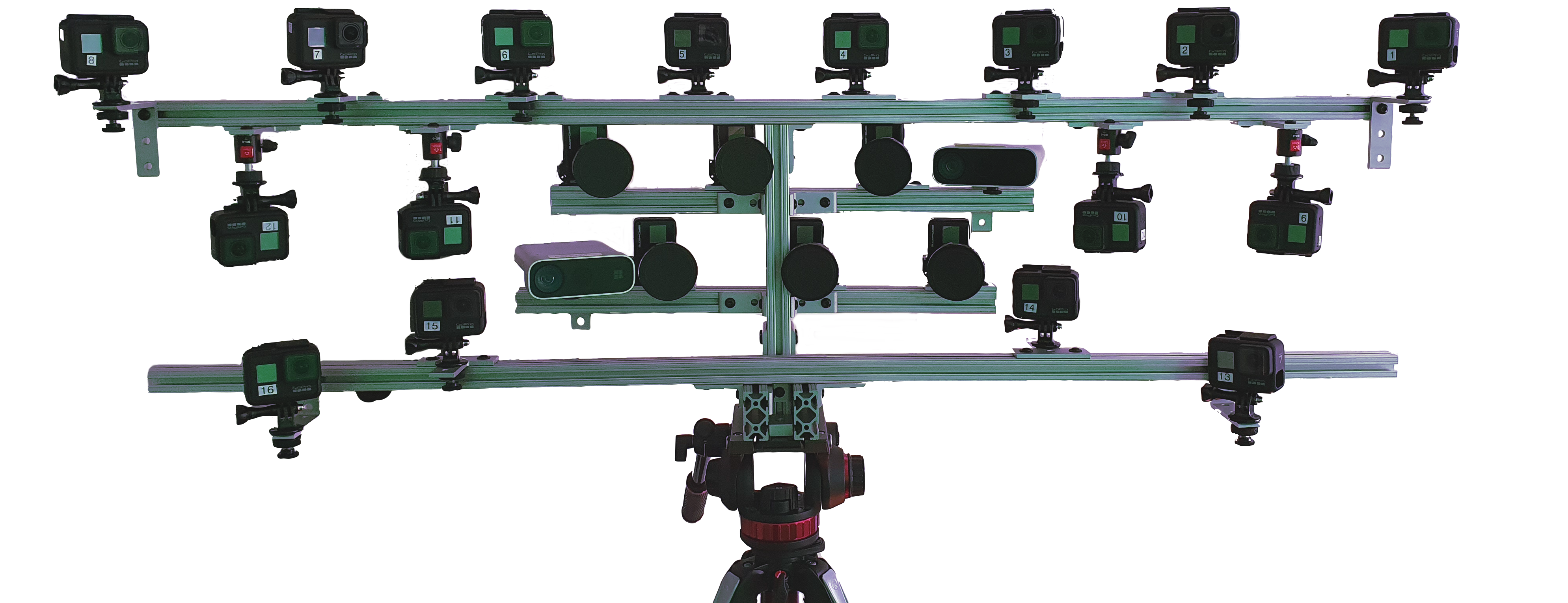}
	\caption{Our custom camera array comprising 16 GoPro HERO7 and 6 \frev{central} Back-Bone H7PRO cameras \frev{(large circular lenses)}. 
	}
	\label{fig:camera_array}
\end{figure}

\paragraph{Human Head Video Dataset} Our dataset consists of 7 multiview captures showing a person performing facial expressions.

\paragraph{Camera Array} Our custom camera array that was used to capture the dataset is comprised of 16 GoPro HERO7 action cameras and 6 Back-Bone H7PRO cameras. 
The Back-Bone cameras are modified GoPro cameras that can fit a standard C-Mount lens. 
Compared to the unmodified GoPro cameras, the Back-Bone cameras have a narrower field-of-view (FoV) and are thus able to capture the subject in more detail. 
We capture at 4k / 30~fps in portrait orientation with the Back-Bone cameras and at 1080p / 60~fps in landscape orientation with the GoPro cameras.
Figure \ref{fig:camera_array} shows a frontal view of the camera array with the six Back-Bone cameras in the center of the array and the GoPro cameras placed around them. 
We capture our subjects from 60\,cm distance and cover approximately 100$^{\circ}$. 

\paragraph{Synchronization} We trigger the camera shutter with a WiFi remote. The cameras do not support a generator lock, so during capture they are only loosely synchronized. We always capture videos for our dataset, even in the cases in which we only use a static frame.
To improve synchronization, we flash an ArUco marker \cite{garrido2014automatic} on a cellphone before each take. 
We then detect the first frame that sees the marker in each video which allows us to synchronize the cameras with an accuracy of 1 frame or better.

\section{Experiments}
\label{sec:experiments}

In this section we show that our method is able to achieve state-of-the-art image reconstruction quality on-par with volumetric methods such as NeRF \cite{mildenhall2020nerf} while allowing for efficient surface reconstruction utilized for real-time rendering in Sec.~\ref{sec:realtimerendering}.

\paragraph{Baselines} 
We compare our method to novel view synthesis techniques with various scene representations. Specifically, we compare to the traditional multi-view stereo of Colmap~\cite{schoenberger2016mvs}, the explicit volumetric representation of Neural Volumes (NV)~\cite{Lombardi:2019}, the implicit volume representation of NeRF~\cite{mildenhall2020nerf}, and the implicit signed distance function of IDR~\cite{yariv2020multiview}. Please refer to the supplemental material for implementation details.

\begin{table*}[]
\footnotesize
\centering
\useunder{\uline}{\ul}{}
\addtolength{\tabcolsep}{-4.5pt}

\begin{tabular}{@{}lrrrrrrrrrrrrrrrrrrr@{}}
\toprule
\multicolumn{1}{c}{\multirow{2}{*}{Scan}} &
  \multicolumn{4}{c}{Colmap~\cite{schoenberger2016mvs}} &
  \multicolumn{4}{c}{IDR~\cite{yariv2020multiview}} &
  \multicolumn{4}{c}{NeRF~\cite{Mildenhall:2019}} &
  \multicolumn{4}{c}{NLR-ST} &
  \multicolumn{3}{c}{NLR-RAS} \\ \cmidrule(l){2-20} 
\multicolumn{1}{c}{} &
  \multicolumn{1}{c}{CD$\downarrow$} &
  \multicolumn{1}{c}{PSNR$\uparrow$} &
  \multicolumn{1}{c}{SSIM$\uparrow$} &
  \multicolumn{1}{c}{LPIPS$\downarrow$} &
  \multicolumn{1}{c}{CD$\downarrow$} &
  \multicolumn{1}{c}{PSNR$\uparrow$} &
  \multicolumn{1}{c}{SSIM$\uparrow$} &
  \multicolumn{1}{c}{LPIPS$\downarrow$} &
  \multicolumn{1}{c}{CD$\downarrow$} &
  \multicolumn{1}{c}{PSNR$\uparrow$} &
  \multicolumn{1}{c}{SSIM$\uparrow$} &
  \multicolumn{1}{c}{LPIPS$\downarrow$} &
  \multicolumn{1}{c}{CD$\downarrow$} &
  \multicolumn{1}{c}{PSNR$\uparrow$} &
  \multicolumn{1}{c}{SSIM$\uparrow$} &
  \multicolumn{1}{c}{LPIPS$\downarrow$} &
  \multicolumn{1}{c}{PSNR$\uparrow$} &
  \multicolumn{1}{c}{SSIM$\uparrow$} &
  \multicolumn{1}{c}{LPIPS$\downarrow$} \\ \midrule
\multicolumn{1}{l|}{65} &
  2.42 &
  15.25 &
  0.862 &
  \multicolumn{1}{r|}{0.143} &
  \textbf{0.70} &
  23.87 &
  0.948 &
  \multicolumn{1}{r|}{0.094} &
  {\ul \textbf{1.02}} &
  \textbf{33.57} &
  \textbf{0.962} &
  \multicolumn{1}{r|}{\textbf{0.055}} &
  {\ul 1.03} &
  {\ul 32.13} &
  {\ul 0.961} &
  \multicolumn{1}{r|}{{\ul 0.063}} &
  31.46 &
  0.960 &
  0.066 \\
\multicolumn{1}{l|}{97} &
  \textbf{0.72} &
  11.93 &
  0.843 &
  \multicolumn{1}{r|}{0.154} &
  {\ul 1.09} &
  23.02 &
  0.921 &
  \multicolumn{1}{r|}{0.117} &
  1.43 &
  28.28 &
  0.933 &
  \multicolumn{1}{r|}{{\ul 0.089}} &
  \textbf{1.16} &
  \textbf{28.48} &
  \textbf{0.939} &
  \multicolumn{1}{r|}{\textbf{0.088}} &
  {\ul 28.37} &
  {\ul 0.939} &
  {\ul 0.089} \\
\multicolumn{1}{l|}{106} &
  {\ul 0.78} &
  15.75 &
  0.887 &
  \multicolumn{1}{r|}{0.141} &
  \textbf{0.58} &
  20.97 &
  0.907 &
  \multicolumn{1}{r|}{0.183} &
  \textbf{0.84} &
  \textbf{33.50} &
  0.947 &
  \multicolumn{1}{r|}{0.092} &
  {\ul 0.82} &
  {\ul 32.98} &
  \textbf{0.951} &
  \multicolumn{1}{r|}{\textbf{0.083}} &
  31.32 &
  {\ul 0.951} &
  {\ul 0.090} \\
\multicolumn{1}{l|}{118} &
  \textbf{0.44} &
  22.73 &
  0.921 &
  \multicolumn{1}{r|}{0.091} &
  {\ul 0.50} &
  22.62 &
  0.944 &
  \multicolumn{1}{r|}{0.139} &
  \textbf{0.88} &
  \textbf{35.62} &
  \textbf{0.966} &
  \multicolumn{1}{r|}{\textbf{0.070}} &
  {\ul 1.71} &
  {\ul 34.87} &
  {\ul 0.964} &
  \multicolumn{1}{r|}{{\ul 0.075}} &
  33.33 &
  0.963 &
  0.078 \\ \bottomrule
\end{tabular}

\caption{\frev{Image error metrics PNSR, SSIM~\cite{wang2004image} and LPIPS~\cite{zhang2018unreasonable} computed on DTU \cite{jensen2014large} for the supervised views. See Supp. for metrics of the held-out views.}
The Chamfer distance \frev{(CD)} is computed based on the scripts from \cite{jensen2014large}. Best scores in bold, second best underlined.
}
  \label{tbl:dtu}
\end{table*}

\begin{figure*}[t!]
  \centering
  \includegraphics[width=\textwidth]{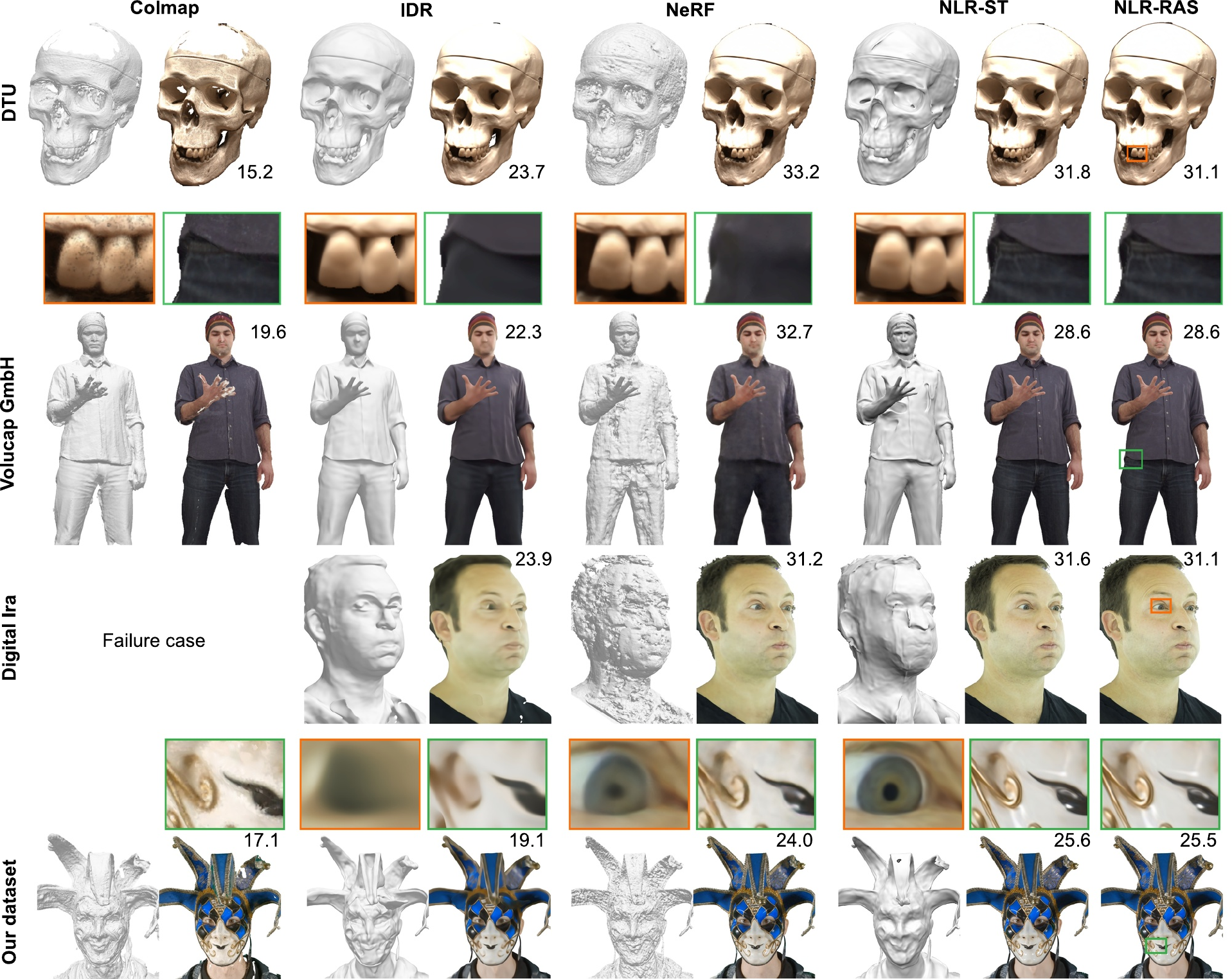}
  \caption{Reconstructed shape and image from various multi-view datasets. \frev{The reprojection error is listed in dB of PSNR.}}
  \label{fig:results}
\end{figure*}

\begin{figure*}[tp!]
  \centering
  \includegraphics[width=\textwidth]{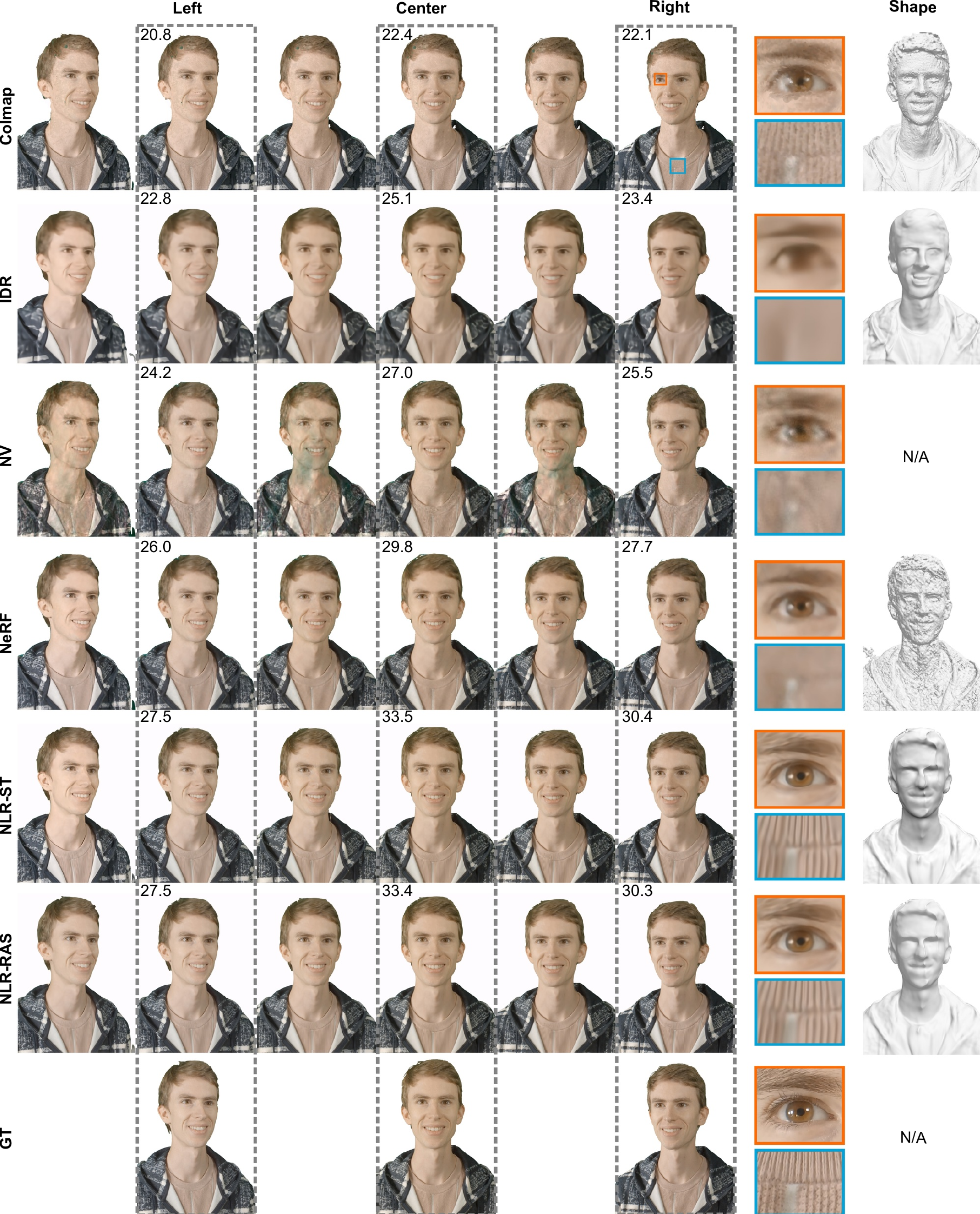}
  \caption{A qualitative comparison for a frame in our dataset. Three supervised views (dashed outline) along with three interpolated views and learned shape. Close-ups enhance detail of the rightmost view. PSNR showed for supervised views.  
  }
  \label{fig:views}
\end{figure*}

\begin{figure*}[h]
  \centering
  \noindent
  \begin{minipage}{0.48\textwidth}
     \centering
  	\includegraphics[width=\textwidth]{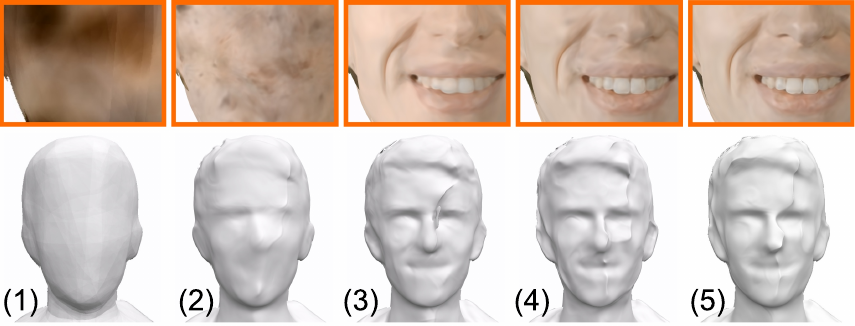}
  \end{minipage}
  \begin{minipage}{0.51\textwidth}    
\footnotesize
\centering
\useunder{\uline}{\ul}{}
\begin{tabular}{@{}lccccrrrr@{}}
\toprule
\multirow{2}{*}{} &
  \multirow{2}{*}{Act.} &
  \multirow{2}{*}{$\mathcal{L}_S$} &
  \multirow{2}{*}{\begin{tabular}[c]{@{}c@{}}Batch\\ size\end{tabular}} &
  \multirow{2}{*}{\begin{tabular}[c]{@{}c@{}}Extra\\ FF\end{tabular}} &
  \multicolumn{2}{c}{PSNR$\uparrow$} &
  \multicolumn{2}{c}{LPIPS$\downarrow$~\cite{zhang2018unreasonable}} \\ \cmidrule(l){6-9} 
 &
   &
   &
   &
   &
  \multicolumn{1}{c}{Train} &
  \multicolumn{1}{c}{Test} &
  \multicolumn{1}{c}{Train} &
  \multicolumn{1}{c}{Test} \\ \midrule
(1) &
  Relu &
  No &
  2K &
  No &
  24.99 &
  10.57 &
  0.171 &
  0.285 \\
(2) &
  Sine &
  No &
  2K &
  No &
  28.49 &
  19.27 &
  0.161 &
  0.239 \\
(3) &
  Sine &
  Yes &
  2K &
  No &
  28.26 &
  24.77 &
  0.151 &
  0.179 \\
(4) &
  Sine &
  Yes &
  50K &
  No &
  30.06 &
  25.20 &
  0.143 &
  0.172 \\
\textbf{(5)} &
  \textbf{Sine} &
  \textbf{Yes} &
  \textbf{50K} &
  \textbf{Yes} &
  \textbf{30.75} &
  \textbf{26.04} &
  \textbf{0.132} &
  \textbf{0.151} \\ \bottomrule
\end{tabular}
  \end{minipage}  
  \caption{
  Learned shapes with interpolated close-ups (left) and \frev{reconstruction metrics} for each condition (right) in our ablation study.
  }
  \label{fig:ablation}
\end{figure*}

\paragraph{Regression performance}
We have used the popular DTU MVS dataset \cite{jensen2014large} with 49 or 64 calibrated camera images along with object masks provided by previous work \cite{yariv2020multiview,Niemeyer2020CVPR} to measure the image reconstruction error \frev{metrics}. We held out three views for testing.
Table~\ref{tbl:dtu} shows that our method achieves SOTA training error comparable with NeRF.
Moreover, our image quality is significantly better than that of our closest competitor, IDR.
We attribute this major separation to the unparalleled representation capacity of {\siren}s.
A qualitative comparison is available in Fig.~\ref{fig:results}.

Additionally, we report the shape reconstruction error as Chamfer distance from the ground-truth provided in the dataset.
Although the shape reconstruction is not our explicit goal, we note that our error is on par with other techniques, though worse than IDR which explicitly focuses on this problem~\cite{yariv2020multiview}. 
We observe that this emerges as a trade-off between the accuracy of view-dependent and high-frequency details in the image reconstruction on one hand, and the view consistency reflected in the geometry on the other one.

\paragraph{View interpolation}
Our angular smoothness loss $\mathcal{L}_S$ is specifically designed to avoid collapse of the emissivity function $\brdf$ for interpolated views. 
We tested its efficiency quantitatively by measuring the image reconstruction error on \frev{test views the held out from results in Table~\ref{tbl:dtu}.
Please refer to the supplement for details.}
As expected, there is a measurable quality drop when compared to the training views observed consistently for all of the methods.
However, the interpolated views produced by our method maintain many of the favorable characteristics from the regression case.
We provide a qualitative comparison of both supervised and interpolated views in Fig.~\ref{fig:views} \frev{and in our video}.

\begin{table}[]
\footnotesize
\centering
\useunder{\uline}{\ul}{}
\addtolength{\tabcolsep}{-4pt}

\begin{tabular}{@{}lllllll@{}}
\toprule
Dataset &
  \multicolumn{1}{c}{Colmap} &
  \multicolumn{1}{c}{IDR} &
  \multicolumn{1}{c}{NV} &
  \multicolumn{1}{c}{NeRF} &
  \multicolumn{1}{c}{NLR-ST} &
  \multicolumn{1}{c}{N.-RAS} \\ \midrule
Volucap (1)    & 19.6/.037 & 22.3/.043 & {\ul 29.3}/.034 & \textbf{32.7}/{\ul .026} & 28.6/\textbf{.022}          & 28.6/\textbf{.022}       \\
Dig.~Ira (1) & Fail.     & 23.9/.286 & 26.5/.287       & {\ul 31.2}/.267    & \textbf{31.6}/\textbf{.255} & 31.1/{\ul .260}       \\
Ours (7)       & 17.6/.187 & 22.3/.202 & \textit{25.1/.186}       & 28.5/.171          & \textbf{30.5}/\textbf{.147} & {\ul 30.3}/{\ul .151} \\ \bottomrule
\end{tabular}

\caption{\frev{Average reconstruction PSNR/LPIPS \cite{zhang2018unreasonable} scores} computed across datasets (number of scenes in parentheses).
\frev{\textit{Italic:}} Only 5 scenes tested.
\frev{See Supplement for an extended version.}
}
  \label{tbl:others}
\end{table}

\paragraph{Human representation}
View-synthesis of human subjects is particularly challenging due to the complex reflection properties of skin, eyes and hair, as well as a lack of high-quality multi-view data.
We address the first challenge with our high-capacity representation network and the latter with our own dataset described in Sec.~\ref{sec:cameraarray}. 
Additionally, we provide experimental results for 360 degree human captures provided by Volucap GmbH~\cite{volucap2020} and high-resolution face captures from the Digital Ira project~\cite{fyffe2014scu}.
Refer to the supplemental for a detailed description of these data.
Table~\ref{tbl:others} summarizes the reconstruction errors and Fig.~\ref{fig:results} shows a few example scenes.
Similar trends as in the DTU datasets can again be observed.
Interestingly, our method achieves a bigger advantage for very high-resolution (3000 $\times$ 4000 px) detailed images in our own dataset.
We speculate that this shows that the traditional ReLU based networks used by IDR and NeRF have reached their capacity, while the explicit representations of Colmap and NV lack easy scaling.
Once again, this does not come at cost of interpolation properties as shown in Figure~\ref{fig:views} and our videos.

\paragraph{Ablation study}
Finally, we verify that the performance of our method is based on the choice of our representation and training procedure.
In Figure and Table~\ref{fig:ablation},  we compare several variants of our method on the scene in Fig.~\ref{fig:views}.
A standard MLP with ReLU does not have the capacity to train a detailed representation (1).
\siren{} remedies this but tends to quickly overfit to the trained pixels (2).
We resolve this first by adding our angular smoothness loss $\mathcal{L}_S$ that regularizes behavior in the angular domain (3), and then by increasing the batch size in order to achieve spatially uniform image quality (4).
Additional Fourier Features \cite{mildenhall2020nerf} for the ray direction remove low frequency noise in $\brdf$ (5).
	
%\petr{Optional: an ablation that studies the role of brdflin continously also with shape chamfer on subsampled DTU.}

%- diagram similar as Lipman \ref{fig:model}
%- table and figure, evaluation on DTU: NeRF, IDF, ours (neural), ours (real-time, optional), colmap, \ref{tbl:dtu} and \ref{fig:results}
%- figure, results without geometry (other datasets) (qualitative with PSNR): NeRF, IDF, ours (neural), ours (real-time, optional), colmap, neural volumes \ref{tbl:others} and \ref{fig:results}
%	- left to right bunch of views, methods vertically, some supervised, some not \ref{fig:views}

\section{Discussion}
\label{sec:discussion}
In summary, we propose a neural rendering framework that optimizes an SDF-based implicit neural scene representation given a set of multi-view images. This framework is unique in combining a representation network architecture using periodic activations with a sphere-tracing-based neural renderer that estimates the shape and view-dependent appearance of the scene. Enabled by a novel loss function that is applied during training, our framework achieves a very high image quality that is comparable with state-of-the-art novel view synthesis methods. As opposed to those methods, our neural representation can be directly converted into a mesh with view-dependent textures that enable high-quality 3D image synthesis in real time using traditional graphics pipelines.

Our approach is not without limitations. Currently, we only consider emissive \frev{radiance} functions that are adequate to model a scene under fixed lighting conditions. Future work could additionally consider dynamic lighting and shading, which some recent neural rendering approaches have started to incorporate~\cite{Meka:2019,zhang2020neural}. 
\frev{Further, similar to IDR~\cite{yariv2020multiview}, our method requires annotated object masks. Automatic image segmentation could be explored in the future to address this.}
Although the synthesized image quality of our approach is competitive with the state of the art, the proxy shapes produced by our method are not quite as accurate as alternative approaches~\cite{schoenberger2016mvs,yariv2020multiview}. While this is not important for the novel view synthesis application we consider in this paper, other applications may benefit from estimating more accurate shapes.
\frev{This includes occasionally visible seam artifacts caused by inaccuracies of the camera calibration.}
Similar to some other recent neural rendering pipelines, ours focuses on overfitting a neural representation on a single 3D scene. An interesting avenue of future work includes the learning of shape spaces, or priors, for certain types of objects, such as faces. While several methods have explored related strategies using conditioning-by-concatenation~\cite{park2019deepsdf,mescheder2019occupancy}, hypernetwork~\cite{sitzmann2019srns}, or meta-learning~\cite{sitzmann2020metasdf} approaches using synthetic data, there is a lack of publicly available photorealistic multi-view image data. In hope of mitigating this shortcoming, \frev{we released our datasets on the project website}. Still, these data may be insufficiently large for learning priors. Finally, although the inference time of our method is fast, the training time is still slow. This hurdle along with the limited computational resources at our disposal is the primary reason preventing us from exploring dynamic video sequences.

\paragraph{Conclusion} Emerging neural rendering approaches are starting to outperform traditional vision and graphics approaches. Yet, traditional graphics pipelines still offer significant practical benefits, such as real-time rendering rates, over these neural approaches. With our work, we take a significant step towards closing this gap, which we believe to be a critical aspect for making neural rendering practical.

\FloatBarrier

\beginacknowledgements
\section*{Acknowledgments}
We would like to thank Volucap GmbH for providing a multi-view video captured using their studio setup as well USC Institute for Creative Technologies for providing a sample of the Digital Ira dataset.

{\small
\bibliographystyle{ieee_fullname}
\bibliography{main}
}

\beginsupplement
\clearpage

\def\refTabDTU{\ref{tbl:dtu}}
\def\refTabOthers{\ref{tbl:others}}
\def\refSecRasEval{\ref{sec:ras_eval}}
\def\refSecCamArray{\ref{sec:cameraarray}}

\section{Real-time rendering analysis}
\label{sec:cg_analysis}

In Section~\refSecRasEval{} of the paper, we compared performance of our real-time renderer to the neural renderer.
Here, we complement this comparison by demonstrating that the rendering quality is a joint product of both the shape representation in $\sdf$ and the emissivity function $\brdf$.
To that goal, we use our new head image dataset and evaluate how well a novel view can be interpolated and/or extrapolated with different choices of geometry and textures.

\paragraph{Setup}
We compare our real-time \frev{rasterized renderer (RAS)} using geometries reconstructed by the surface based methods of Colmap~\cite{schoenberger2016mvs}, IDR~\cite{yariv2020multiview} and \frev{our Neural Lumigraph Rendering (NLR)}.
Additionally, we consider both the neural textures generated by the respective method (not available for Colmap) as well as alternative usage of the original training camera images.

We provide the renderer with textures corresponding to 5 of the 6 high-resolution central Back-Bone H7PRO cameras in our dataset (see Section~\refSecCamArray{}), and we measure the PSNR of the held-out interpolated/extrapolated 6$^{\mathrm{th}}$ view. The same ground-truth masks are applied for all measurements. We average results from all 6 possible test view choices.

% \usepackage{booktabs}
% \usepackage{multirow}
\begin{table}[]
\footnotesize
\centering
\useunder{\uline}{\ul}{}
\begin{tabular}{@{}llrrrr@{}}
\toprule
\multicolumn{1}{c}{} &
  \multicolumn{1}{c}{} &
  \multicolumn{1}{c}{} &
  \multicolumn{3}{c}{Neural textures} \\ \cmidrule(l){4-6} 
\multicolumn{1}{c}{\multirow{-2}{*}{Scene}} &
  \multicolumn{1}{c}{\multirow{-2}{*}{Method}} &
  \multicolumn{1}{c}{\multirow{-2}{*}{\begin{tabular}[c]{@{}c@{}}Captured\\ textures\end{tabular}}} &
  \multicolumn{1}{c}{1$\times$} &
  \multicolumn{1}{c}{2$\times$} &
  \multicolumn{1}{c}{3$\times$} \\ \midrule
 &
  Colmap &
  \cellcolor[HTML]{E8F4EE}\textbf{23.96} &
  N/A &
  N/A &
  N/A \\
 &
  IDR &
  \cellcolor[HTML]{FBF7FA}22.23 &
  \cellcolor[HTML]{FBFAFD}22.49 &
  \cellcolor[HTML]{F3F9F7}23.31 &
  \cellcolor[HTML]{EEF7F3}23.61 \\
\multirow{-3}{*}{A1} &
  \frev{NLR (RAS)} &
  \cellcolor[HTML]{EDF6F2}23.65 &
  \cellcolor[HTML]{EAF5F0}\textbf{23.83} &
  \cellcolor[HTML]{B8E1C4}\textbf{26.93} &
  \cellcolor[HTML]{80CA94}\textbf{30.33} \\
 &
  Colmap &
  \cellcolor[HTML]{F99C9E}11.48 &
  N/A &
  N/A &
  N/A \\
 &
  IDR &
  \cellcolor[HTML]{FBE3E6}19.80 &
  \cellcolor[HTML]{FBE1E4}19.65 &
  \cellcolor[HTML]{FBEBEE}20.80 &
  \cellcolor[HTML]{FBEEF1}21.15 \\
\multirow{-3}{*}{A8} &
  \frev{NLR (RAS)} &
  \cellcolor[HTML]{F9FBFD}\textbf{22.90} &
  \cellcolor[HTML]{F7FAFB}\textbf{23.06} &
  \cellcolor[HTML]{C7E7D1}\textbf{26.00} &
  \cellcolor[HTML]{93D2A4}\textbf{29.20} \\
 &
  Colmap &
  \cellcolor[HTML]{F8696B}5.45 &
  N/A &
  N/A &
  N/A \\
 &
  IDR &
  \cellcolor[HTML]{FBD8DB}18.55 &
  \cellcolor[HTML]{FBD8DA}18.49 &
  \cellcolor[HTML]{FBDBDE}18.89 &
  \cellcolor[HTML]{FBDCDF}18.99 \\
\multirow{-3}{*}{L1} &
  \frev{NLR (RAS)} &
  \cellcolor[HTML]{FBEFF2}\textbf{21.24} &
  \cellcolor[HTML]{FBF2F5}\textbf{21.61} &
  \cellcolor[HTML]{E5F3EB}\textbf{24.13} &
  \cellcolor[HTML]{BEE3C9}\textbf{26.56} \\
 &
  Colmap &
  \cellcolor[HTML]{FBE5E7}20.03 &
  N/A &
  N/A &
  N/A \\
 &
  IDR &
  \cellcolor[HTML]{EFF7F4}23.56 &
  \cellcolor[HTML]{EDF6F2}23.66 &
  \cellcolor[HTML]{D3ECDC}25.24 &
  \cellcolor[HTML]{C8E7D2}25.95 \\
\multirow{-3}{*}{L3} &
  \frev{NLR (RAS)} &
  \cellcolor[HTML]{E2F2E9}\textbf{24.32} &
  \cellcolor[HTML]{DAEEE1}\textbf{24.85} &
  \cellcolor[HTML]{9AD5AA}\textbf{28.76} &
  \cellcolor[HTML]{63BE7B}\textbf{32.10} \\
 &
  Colmap &
  \cellcolor[HTML]{F9ACAE}13.36 &
  N/A &
  N/A &
  N/A \\
 &
  IDR &
  \cellcolor[HTML]{FAD3D6}17.98 &
  \cellcolor[HTML]{FAD1D4}17.73 &
  \cellcolor[HTML]{FAD6D9}18.36 &
  \cellcolor[HTML]{FBD8DB}18.59 \\
\multirow{-3}{*}{L4} &
  \frev{NLR (RAS)} &
  \cellcolor[HTML]{FBE1E4}\textbf{19.57} &
  \cellcolor[HTML]{FBE3E5}\textbf{19.80} &
  \cellcolor[HTML]{FBF7FA}\textbf{22.20} &
  \cellcolor[HTML]{E3F2EA}\textbf{24.25} \\
 &
  Colmap &
  \cellcolor[HTML]{FBD7DA}18.40 &
  N/A &
  N/A &
  N/A \\
 &
  IDR &
  \cellcolor[HTML]{FBF6F9}22.07 &
  \cellcolor[HTML]{FBF4F7}21.84 &
  \cellcolor[HTML]{FBFBFE}22.60 &
  \cellcolor[HTML]{F8FBFC}22.96 \\
\multirow{-3}{*}{M2} &
  \frev{NLR (RAS)} &
  \cellcolor[HTML]{E8F4EE}\textbf{23.94} &
  \cellcolor[HTML]{E2F2E9}\textbf{24.34} &
  \cellcolor[HTML]{B7E0C4}\textbf{26.96} &
  \cellcolor[HTML]{8ACE9C}\textbf{29.76} \\
 &
  Colmap &
  \cellcolor[HTML]{E4F3EA}\textbf{24.22} &
  N/A &
  N/A &
  N/A \\
 &
  IDR &
  \cellcolor[HTML]{FAC3C5}16.06 &
  \cellcolor[HTML]{F9B0B3}13.85 &
  \cellcolor[HTML]{FAB8BA}14.78 &
  \cellcolor[HTML]{FABFC1}15.59 \\
\multirow{-3}{*}{P4} &
  \frev{NLR (RAS)} &
  \cellcolor[HTML]{E5F3EB}24.16 &
  \cellcolor[HTML]{E3F2EA}\textbf{24.27} &
  \cellcolor[HTML]{B0DDBD}\textbf{27.41} &
  \cellcolor[HTML]{85CC98}\textbf{30.04} \\ \midrule
 &
  Colmap &
  \cellcolor[HTML]{FAC8CB}16.70 &
  N/A &
  N/A &
  N/A \\
 &
  IDR &
  \cellcolor[HTML]{FBE5E8}20.04 &
  \cellcolor[HTML]{FBE2E4}19.67 &
  \cellcolor[HTML]{FBE9EC}20.57 &
  \cellcolor[HTML]{FBEDF0}20.98 \\
\multirow{-3}{*}{Average} &
  \frev{NLR (RAS)} &
  \cellcolor[HTML]{FBFCFE}\textbf{22.83} &
  \cellcolor[HTML]{F6FAFA}\textbf{23.11} &
  \cellcolor[HTML]{C6E6D0}\textbf{26.06} &
  \cellcolor[HTML]{98D4A9}\textbf{28.89} \\ \bottomrule
\end{tabular}
\caption{PSNR scores for the interpolated/extrapolated views achieved by different variants of our real-time renderer. Neural textures for the original camera poses (1$\times$), and their supersampled variants (2$\times$, 3$\times$) are compared to a direct use of the original images captured by our camera array.
}
  \label{tbl:cg_res}
\end{table}

\paragraph{Results}

\begin{sidewaysfigure*}[ht!]
  \centering
  \includegraphics[width=\textwidth]{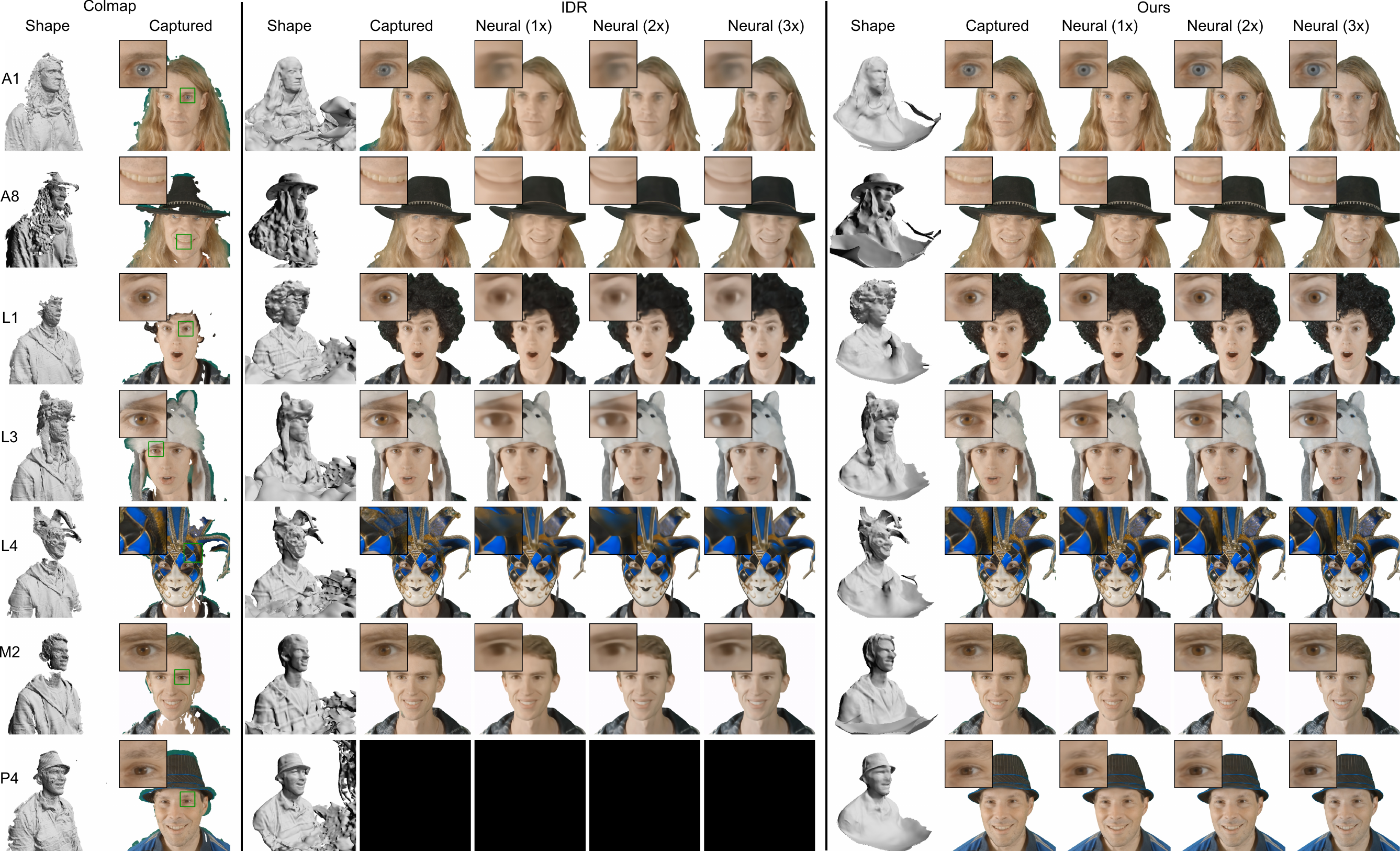}
  \caption{Comparison of our real-time rendering variant evaluated on an view not present in the texture set.
  The IDR results for the presented view in P4 are black due to an occlusion by a false geometry generated in front of the person (see the shape rendering). While fixable and only present in this scene, we keep the geometry as is and provide detailed metrics for each single scene in Table~\ref{tbl:cg_res}. A rejection of this result would not change the overall trend.}
  \label{fig:cg_res}
\end{sidewaysfigure*}

Table~\ref{tbl:cg_res} presents PSNR scores for each scene as well as the overall average.
A qualitative comparison is presented in Fig~\ref{fig:cg_res}.
For all choices of textures, we observe a favorable performance of the shape exported from our method compared to shapes exported by both the IDR and Colmap.
Colmap generates very accurate but incomplete geometries resulting in holes in the rendered images (note the hair in the scene ``L1'').
The geometry extracted from IDR provides similar degree of view consistency as our own geometry when combined with the original captured textures, even though our method still maintains a small margin (see the column ``Captured textures'' in Tab.~\ref{tbl:cg_res}).
However, the neural textures generated by IDR lack detail present in our neural textures which results in comparatively lower PSNR scores (see the column ``Neural 1$\times$'').

\begin{figure}[h]
  \centering
  \includegraphics[width=0.8\linewidth]{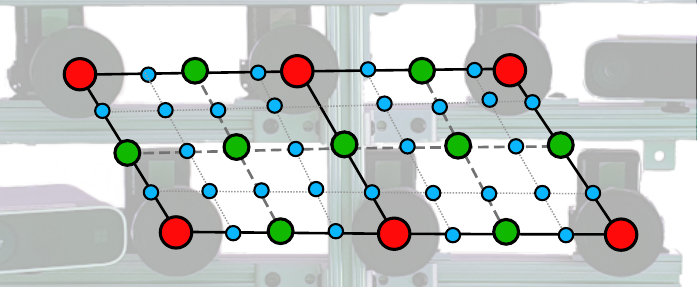}
  \caption{The positions of the original texturing cameras (red) and the interpolation pattern to generate novel textures labeled as ``2$\times$'' (green) and ``3$\times$'' (blue).}
  \label{fig:upsampling}
\end{figure}

\paragraph{Texture space upsampling}

An important argument for using the neural textures rather than the original captured images is the possibility to use the neural renderer to produce additional virtual views.
This allows to generate much denser virtual camera spacing, i.e. through view synthesis, than would be practical with physical camera setups.
To demonstrate this unique capability we resampling the texture space by subdividing the original $3\,\times\,2$ camera layout in our dataset as shown in Figure~\ref{fig:upsampling}.
This yields first the original 6 textures (labeled ``1$\times$''), then additional 9 textures (total of 15; labeled ``2$\times$'') and finally additional 30 textures (total of 45; labeled ``3$\times$'').
We follow the same procedure and remove the texture corresponding to the original ground-truth camera pose for the test.

The results are again presented in Table~\ref{tbl:cg_res} (two right-most columns) and in Fig~\ref{fig:cg_res}.
Both IDR and SineSDF experience significant quality boost as the texturing angular space gets denser and the interpolation distances between blended textures get smaller.
However, only our technique can leverage the high spatial detail featured in our neural textures and, as a result, shows significantly larger PSNR overall.
Please explore the video supplement to evaluate dynamic properties of this behavior.

\section{Additional Results}

\begin{table*}[]
\footnotesize
\centering
\useunder{\uline}{\ul}{}
\addtolength{\tabcolsep}{-2.5pt}

\begin{tabular}{@{}lrrrrrrrrrrrrrrr@{}}
\toprule
\multicolumn{1}{c}{\multirow{2}{*}{Scan}} &
  \multicolumn{3}{c}{Colmap~\cite{schoenberger2016mvs}} &
  \multicolumn{3}{c}{IDR~\cite{yariv2020multiview}} &
  \multicolumn{3}{c}{NeRF~\cite{Mildenhall:2019}} &
  \multicolumn{3}{c}{NLR-ST} &
  \multicolumn{3}{c}{NLR-RAS} \\ \cmidrule(l){2-16} 
\multicolumn{1}{c}{} &
  \multicolumn{1}{c}{PSNR$\uparrow$} &
  \multicolumn{1}{c}{SSIM$\uparrow$} &
  \multicolumn{1}{c}{LPIPS$\downarrow$} &
  \multicolumn{1}{c}{PSNR$\uparrow$} &
  \multicolumn{1}{c}{SSIM$\uparrow$} &
  \multicolumn{1}{c}{LPIPS$\downarrow$} &
  \multicolumn{1}{c}{PSNR$\uparrow$} &
  \multicolumn{1}{c}{SSIM$\uparrow$} &
  \multicolumn{1}{c}{LPIPS$\downarrow$} &
  \multicolumn{1}{c}{PSNR$\uparrow$} &
  \multicolumn{1}{c}{SSIM$\uparrow$} &
  \multicolumn{1}{c}{LPIPS$\downarrow$} &
  \multicolumn{1}{c}{PSNR$\uparrow$} &
  \multicolumn{1}{c}{SSIM$\uparrow$} &
  \multicolumn{1}{c}{LPIPS$\downarrow$} \\ \midrule
65 &
  15.12 &
  0.845 &
  0.156 &
  21.89 &
  0.939 &
  0.105 &
  \textbf{27.08} &
  \textbf{0.948} &
  \textbf{0.069} &
  {\ul 26.58} &
  {\ul 0.941} &
  {\ul 0.083} &
  25.60 &
  0.940 &
  0.087 \\
97 &
  11.37 &
  0.835 &
  0.166 &
  22.77 &
  0.914 &
  0.128 &
  24.17 &
  0.918 &
  {\ul 0.109} &
  \textbf{26.43} &
  \textbf{0.922} &
  \textbf{0.108} &
  {\ul 26.40} &
  {\ul 0.922} &
  {\ul 0.108} \\
106 &
  16.69 &
  0.875 &
  0.153 &
  20.49 &
  0.891 &
  0.205 &
  \textbf{30.17} &
  0.934 &
  0.107 &
  {\ul 29.60} &
  \textbf{0.939} &
  \textbf{0.098} &
  28.52 &
  {\ul 0.938} &
  {\ul 0.104} \\
118 &
  21.72 &
  0.912 &
  0.100 &
  23.24 &
  0.937 &
  0.150 &
  \textbf{31.03} &
  \textbf{0.955} &
  \textbf{0.083} &
  {\ul 30.77} &
  {\ul 0.950} &
  {\ul 0.091} &
  30.39 &
  0.950 &
  0.093 \\ \bottomrule
\end{tabular}

\caption{\frev{Image error metrics PNSR, SSIM~\cite{wang2004image} and LPIPS~\cite{zhang2018unreasonable} computed for the 3 held-out test views of the DTU dataset \cite{jensen2014large} complementing the respective training errors in Table~2 of the main paper. Best scores in bold, second best underlined.
}}
  \label{tbl:dtu_test}
\end{table*}

\begin{table*}[]
\footnotesize
\centering
\useunder{\uline}{\ul}{}
\addtolength{\tabcolsep}{-5pt}

\begin{tabular}{@{}lcrrrrrrrrrrrrrrrrr@{}}
\toprule
\multirow{2}{*}{Dataset} &
  \multicolumn{3}{c}{Colmap} &
  \multicolumn{3}{c}{IDR} &
  \multicolumn{3}{c}{NV} &
  \multicolumn{3}{c}{NeRF} &
  \multicolumn{3}{c}{NLR-ST} &
  \multicolumn{3}{c}{NLR-RAS} \\ \cmidrule(l){2-19} 
 &
  \multicolumn{1}{l}{PSNR$\uparrow$} &
  \multicolumn{1}{l}{SSIM$\uparrow$} &
  \multicolumn{1}{l}{LPIPS$\downarrow$} &
  \multicolumn{1}{l}{PSNR$\uparrow$} &
  \multicolumn{1}{l}{SSIM$\uparrow$} &
  \multicolumn{1}{l}{LPIPS$\downarrow$} &
  \multicolumn{1}{l}{PSNR$\uparrow$} &
  \multicolumn{1}{l}{SSIM$\uparrow$} &
  \multicolumn{1}{l}{LPIPS$\downarrow$} &
  \multicolumn{1}{l}{PSNR$\uparrow$} &
  \multicolumn{1}{l}{SSIM$\uparrow$} &
  \multicolumn{1}{l}{LPIPS$\downarrow$} &
  \multicolumn{1}{l}{PSNR$\uparrow$} &
  \multicolumn{1}{l}{SSIM$\uparrow$} &
  \multicolumn{1}{l}{LPIPS$\downarrow$} &
  \multicolumn{1}{l}{PSNR$\uparrow$} &
  \multicolumn{1}{l}{SSIM$\uparrow$} &
  \multicolumn{1}{l}{LPIPS$\downarrow$} \\ \midrule
Volucap (1) &
  \multicolumn{1}{r}{19.57} &
  0.975 &
  0.037 &
  22.27 &
  0.978 &
  0.043 &
  {\ul 29.32} &
  0.975 &
  0.034 &
  \textbf{32.71} &
  \textbf{0.982} &
  {\ul 0.026} &
  28.59 &
  {\ul 0.981} &
  \textbf{0.022} &
  28.57 &
  {\ul 0.981} &
  \textbf{0.022} \\
Digit.~Ira (1) &
  \multicolumn{3}{c}{Fail.} &
  23.89 &
  0.835 &
  0.286 &
  26.54 &
  0.829 &
  0.287 &
  {\ul 31.18} &
  {\ul 0.845} &
  0.267 &
  \textbf{31.63} &
  \textbf{0.851} &
  \textbf{0.255} &
  31.13 &
  \textbf{0.851} &
  {\ul 0.260} \\
Ours (7) &
  \multicolumn{1}{r}{17.58} &
  0.841 &
  0.187 &
  22.34 &
  0.878 &
  0.202 &
  25.05 &
  0.857 &
  0.186 &
  28.45 &
  0.898 &
  0.171 &
  \textbf{30.45} &
  \textbf{0.921} &
  \textbf{0.147} &
  {\ul 30.25} &
  \textbf{0.921} &
  {\ul 0.151} \\ \bottomrule
\end{tabular}

\caption{\frev{Average image reconstruction error metrics PNSR, SSIM~\cite{wang2004image} and LPIPS~\cite{zhang2018unreasonable} computed across datasets (number of scenes in parentheses).
(*) Only 5 scenes tested.
}}
  \label{tbl:others_full}
\end{table*}

\frev{
In this section, we provide additional results that did not fit into the main paper. 
Table~\ref{tbl:dtu_test} complements Table~\refTabDTU{} in the main paper and shows results of the image reconstruction metrics computed for the three test views held-out from the training on the DTU dataset \cite{jensen2014large}.
Table~\ref{tbl:others_full} extends Table~\refTabOthers{} in the main paper and shows additional image metric scores for results using variety of mutli-view datasets.
}

\section{Supplemental Video}

We provide a detailed video supplement to fully evaluate view consistency of our novel rendered views and compare to the baseline methods.
A summary video with side by side comparison and comments is available in the root folder as {\texttt{4945\_supplemental\_video.mp4}} (x264 encoding in mp4 container). 
Additionally, individual results in form of short videos are provided separately in three folders, each showing a different type of comparison.

\paragraph{Neural renderings}
Videos in the folder {\texttt{neural\_videos}} present neural renderings of scenes from the datasets used in our paper produced by our method as well as well Colmap, IDR, Neural Volumes and NeRF.
Note, that the renderings are provided without post-processing and show the background reconstruction present in Neural Volumes and NeRF.
This background was not evaluated in metrics in our paper; all tested methods use the same foreground mask when evaluating the PSNR.

\paragraph{Shape renderings}
In the folder {\texttt{shape\_videos}}, we use a simple OpenGL renderer using the PyRender Python package to visualize meshes extracted from Colmap, IDR and our \frev{Neural Lumigraph} that are later used for the real-time rendering.
Note that IDR contains lot of boundary geometry noise for views that were not supervised during training.

\paragraph{Real-time renderings}
Finally, in the folder {\texttt{cg\_videos}}, we present outputs of our real-time renderer as well as different variants presented in the study in Sup. Sec.~\ref{sec:cg_analysis}.

\section{Input sensitivity}

\begin{figure}[h]
  \centering
  \includegraphics[width=1.0\linewidth]{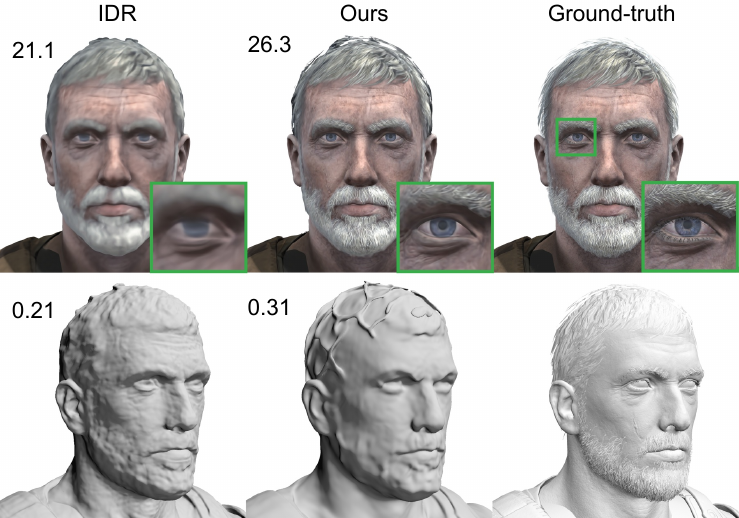}
  \caption{A comparison of image and shape reconstruction by IDR and our method with the ground-truth Unity rendering and mesh shown on the right. The upper numbers denote the image PSNR averaged over all views and the lower numbers correspond to the Chamfer distance.}
  \label{fig:unity}
\end{figure}

Some of the key input properties that affects quality of multi-view reconstruction is the accuracy of camera calibration.
We investigate how our method performs if this factor is taken out of the equation by measuring its performance on a computer-generated 3D scene.
We have used Unity to render a 5\,$\times$\,5 grid of views of a 3D head model with surface specular properties at a resolution of 2160\,$\times$\,2160 pixel and we extracted the ground-truth camera poses as well as the object masks.

As Fig.~\ref{fig:unity} shows, both methods achieve a clean shape reconstructions without artifacts. 
The hair region is represented by a cloud of semi-transparent billboards in the original 3D model, so this would be challenging to be accurately reconstructed by any approach.
While the Chamfer distance metric favors the results of IDR, we see that our image quality is much higher as it features notably higher spatial fidelity.
We conclude that our method can avoid geometry artifacts if consistent camera and image labeling can be obtained.

\section{Datasets}

\paragraph{Our Captured Dataset}
Each capture consists of 22 images and corresponding foreground masks. Six images are captured with high-resolution narrow field-of-view (FOV) cameras, while the remaining 16 images are captured with low-resolution wide FOV cameras.
While these additional 16 images provide useful geometrical information, they do not carry high quality visual detail.
This is why we use all views for training each of the evaluated methods but we only evaluate reconstruction quality for the 6 central high-resolution cameras.

\begin{figure*}[h]
	\newcommand\datasetwidth{0.45\linewidth}
	\newcommand\imagewidth{0.95\linewidth}
	\centering
	\begin{minipage}[t]{\datasetwidth}
		\centering
		\includegraphics[width=\imagewidth]{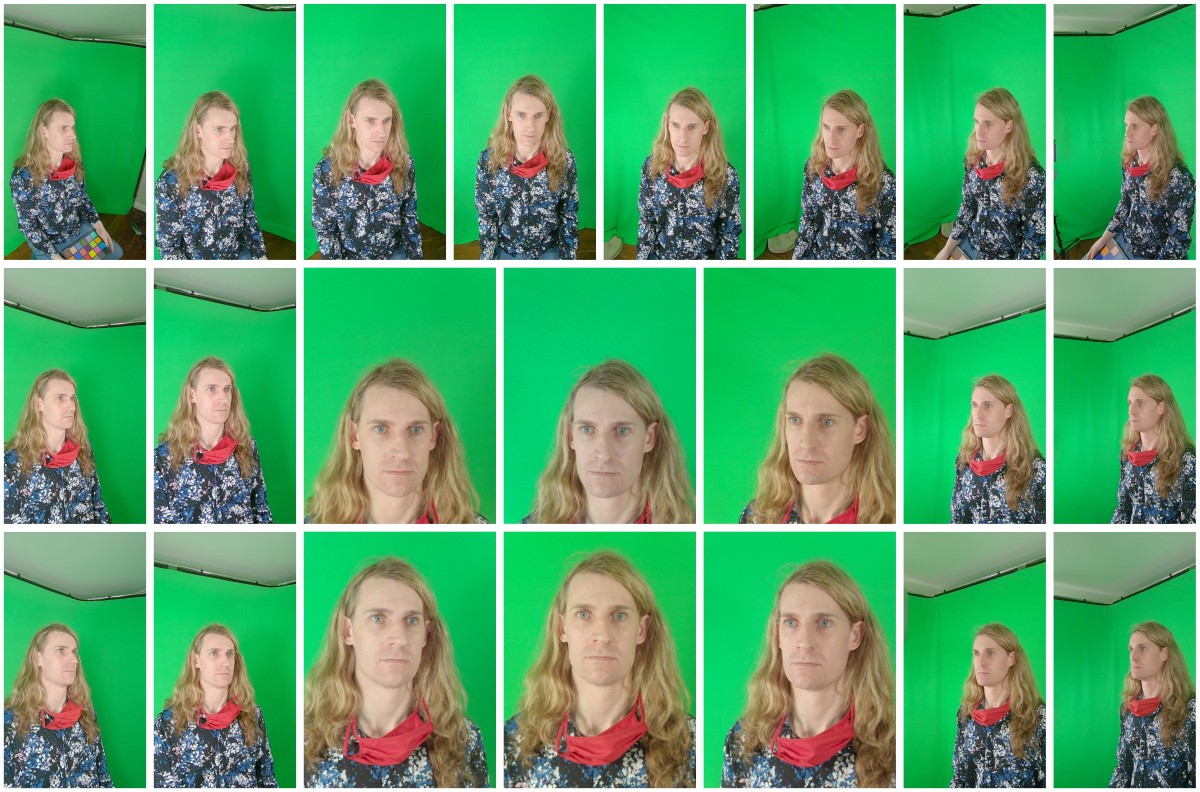}
		\caption{A1}
		\label{subfig:andrew_v1}
	\end{minipage}
	\begin{minipage}[t]{\datasetwidth}
		\centering
		\includegraphics[width=\imagewidth]{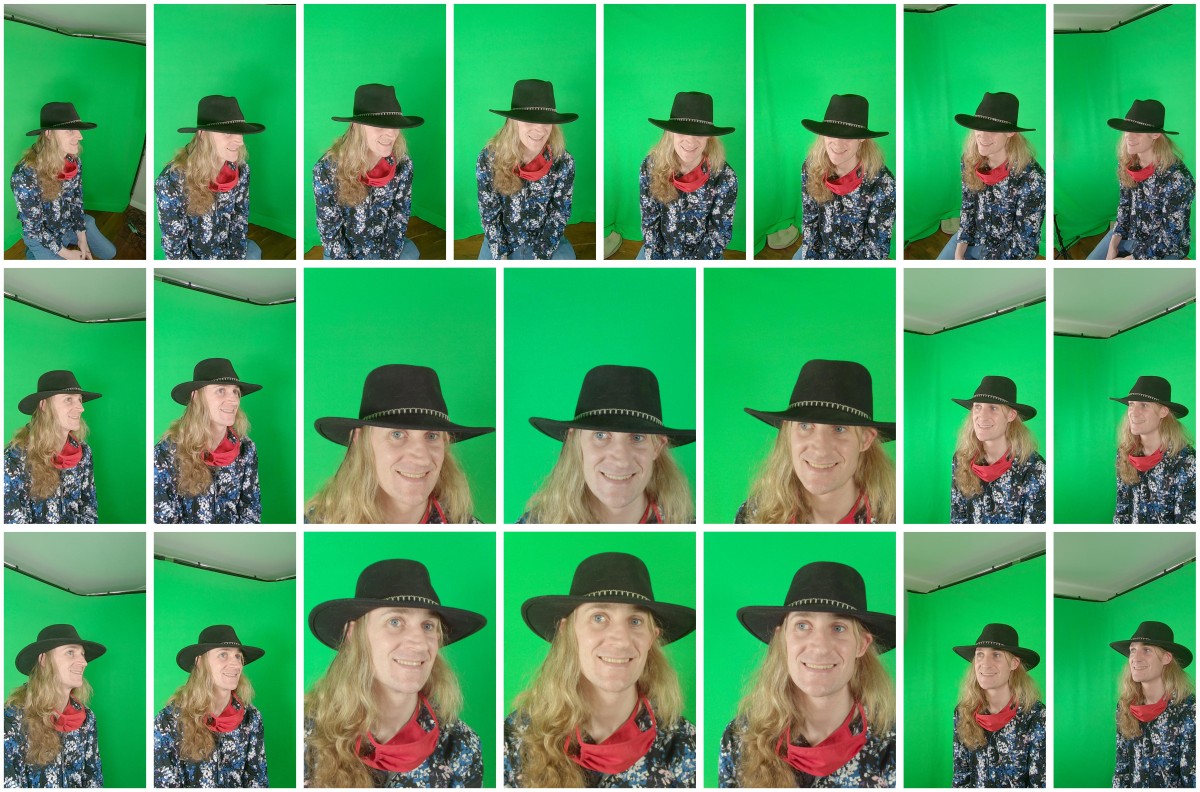}
		\caption{A8}
		\label{subfig:andrew_v8}
	\end{minipage}
	\begin{minipage}[t]{\datasetwidth}
		\centering
		\includegraphics[width=\imagewidth]{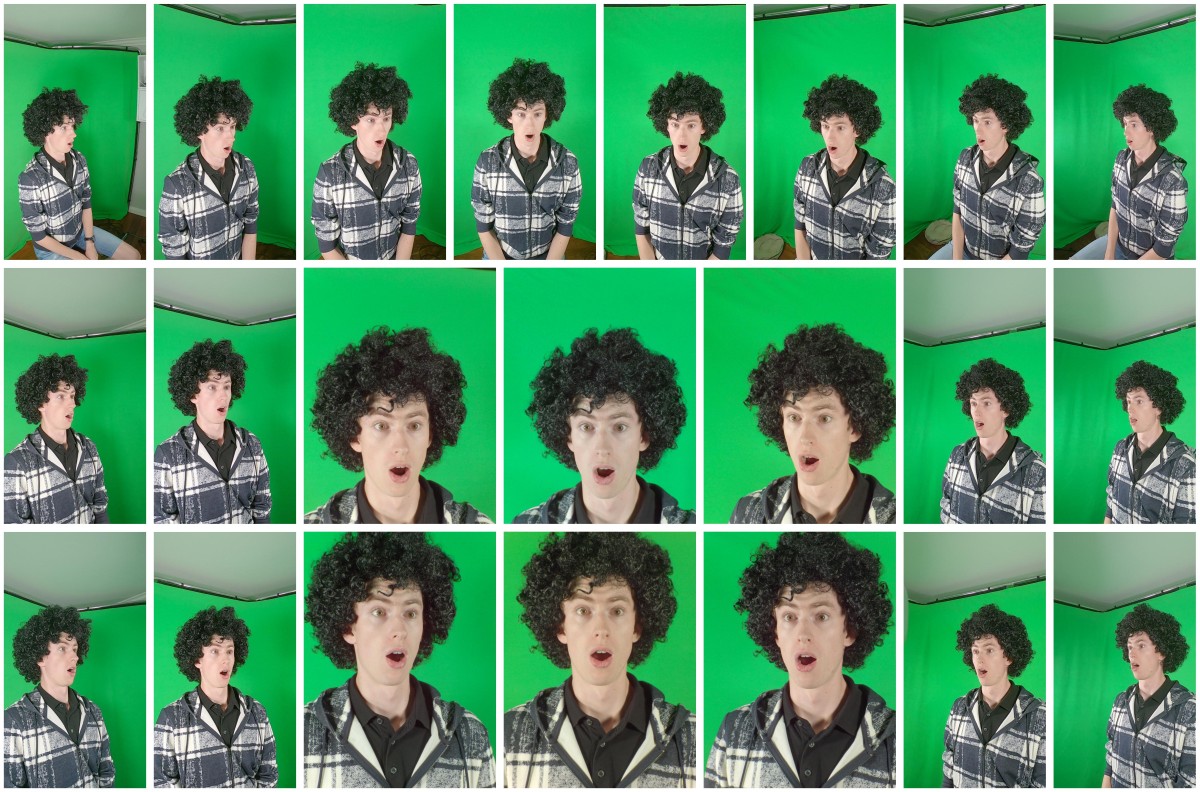}
		\caption{L1}
		\label{fig:lars_v1}
	\end{minipage}
	\begin{minipage}[t]{\datasetwidth}
		\centering
		\includegraphics[width=\imagewidth]{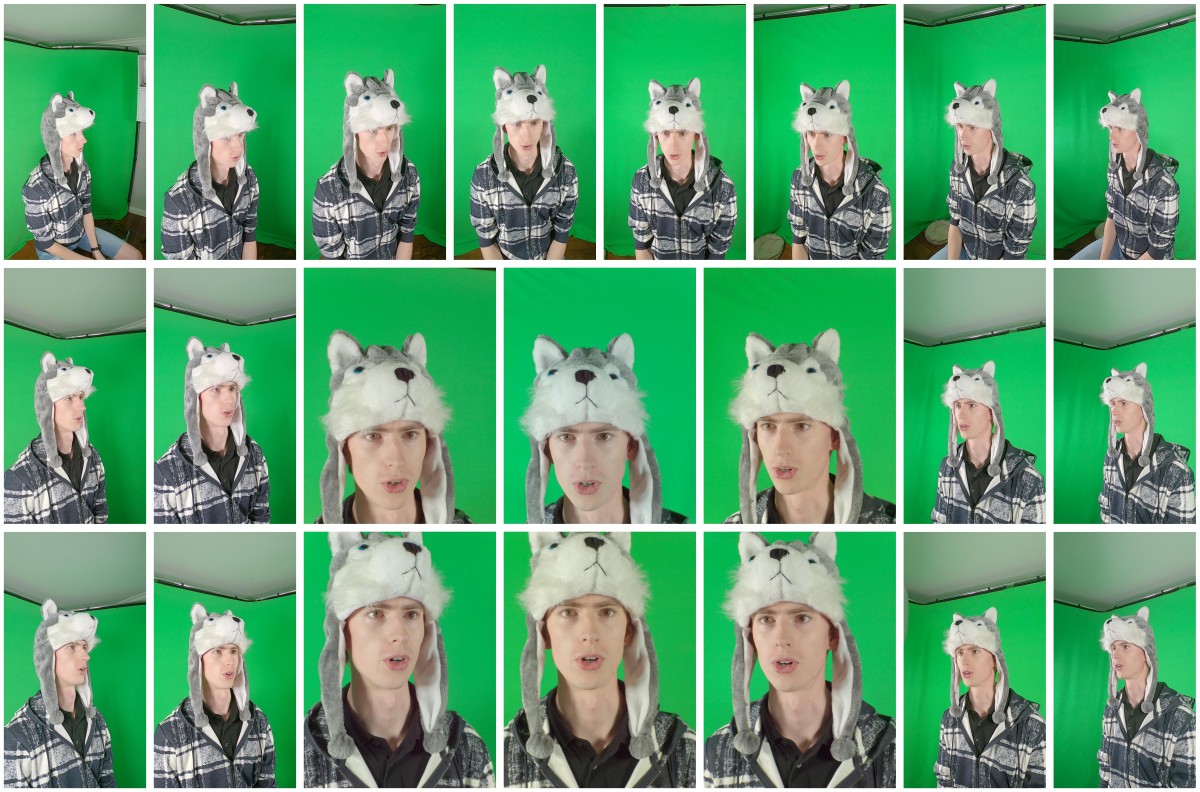}
		\caption{L3}
		\label{subfig:lars_v3}
	\end{minipage}
	\begin{minipage}[t]{\datasetwidth}
		\centering
		\includegraphics[width=\imagewidth]{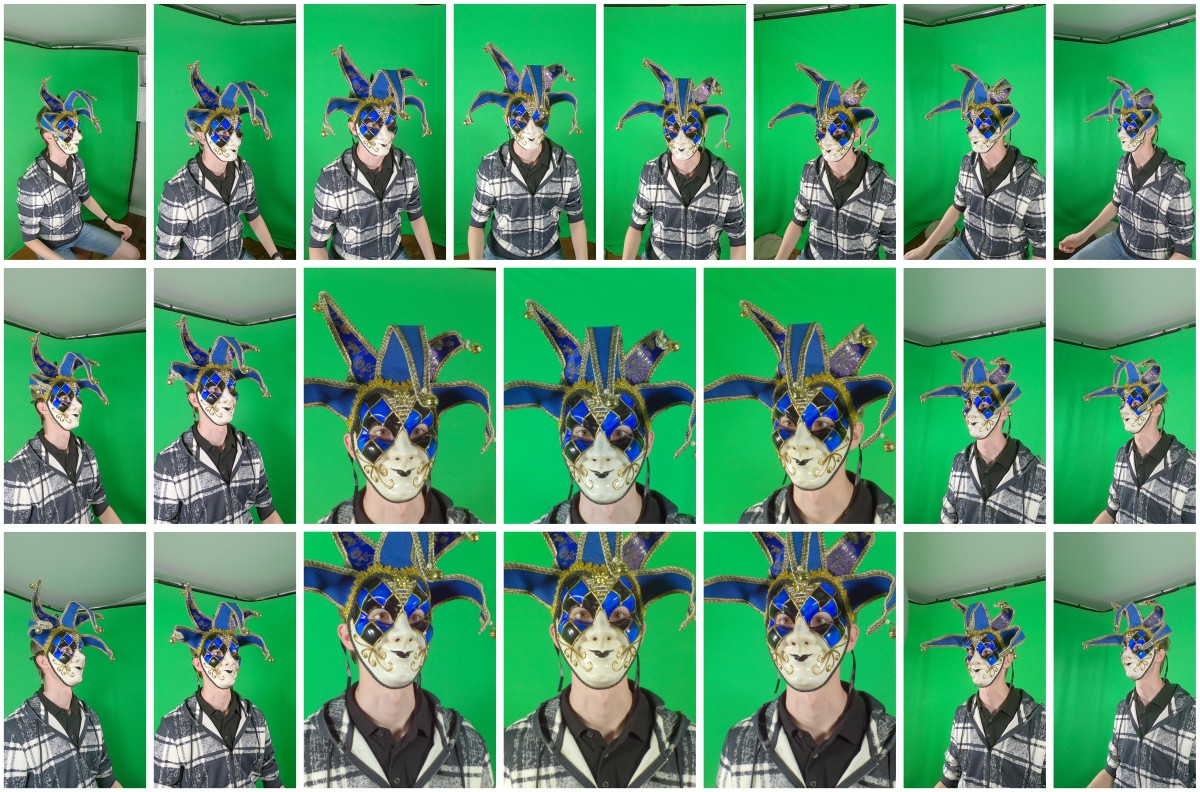}
		\caption{L4}
		\label{subfig:lars_v4}
	\end{minipage}
	\begin{minipage}[t]{\datasetwidth}
		\centering
		\includegraphics[width=\imagewidth]{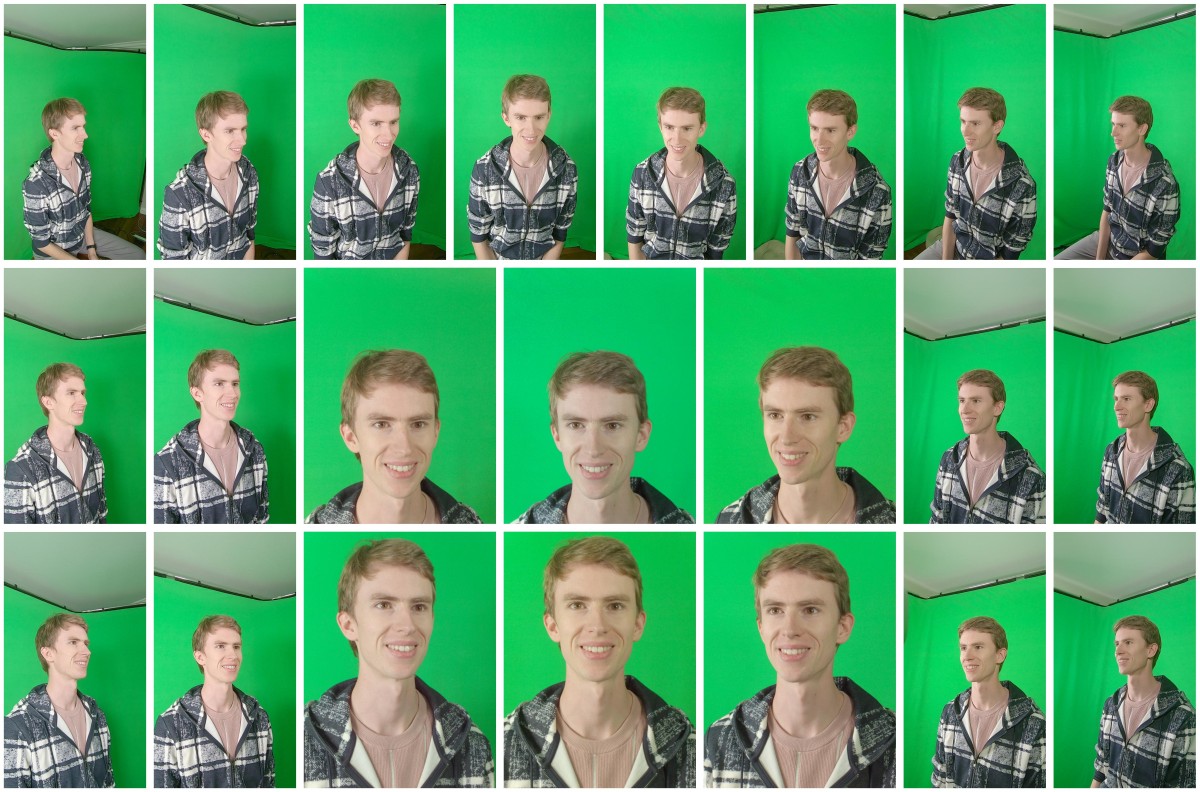}
		\caption{M2}
		\label{subfig:many_cams_v2}
	\end{minipage}
	\begin{minipage}[t]{\datasetwidth}
		\centering
		\includegraphics[width=\imagewidth]{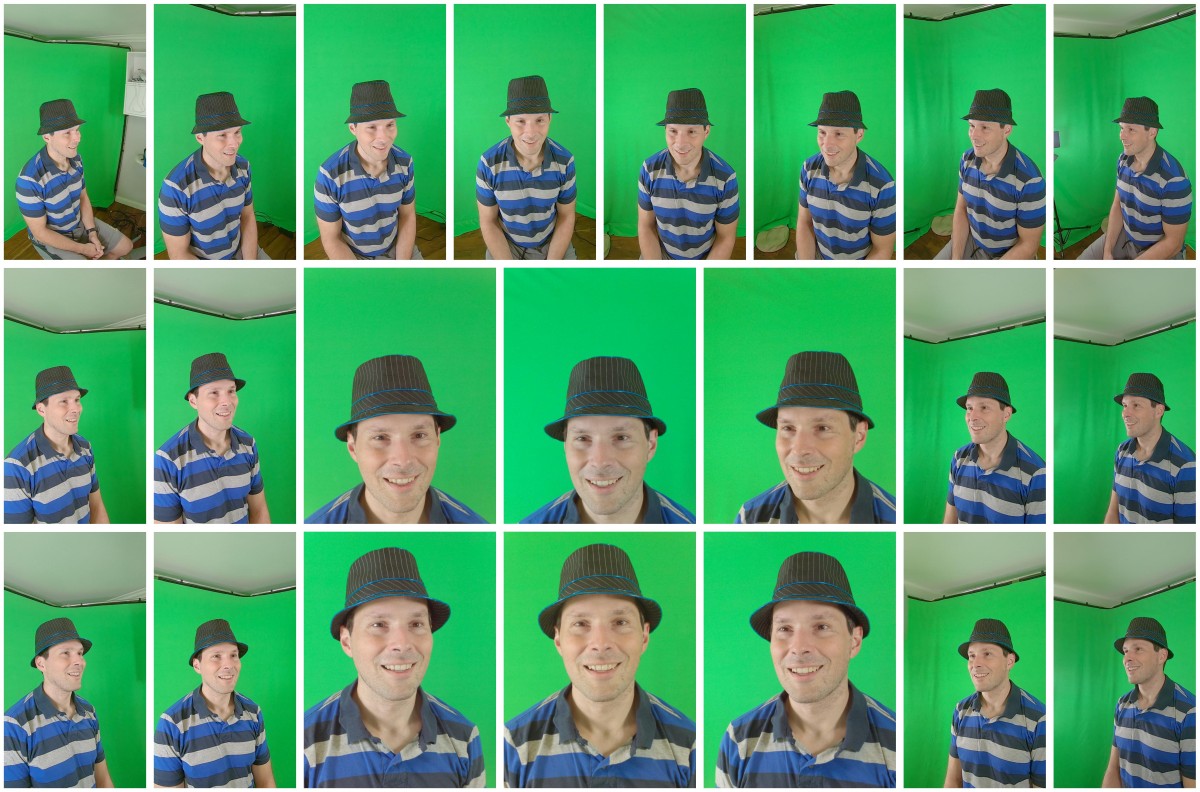}
		\caption{P4}
		\label{subfig:petr_v3}
	\end{minipage}
	\caption{The seven scenes forming our dataset. Image positions approximate the camera layout. The central 6 images are high-resolution head close-ups.}
	\label{fig:our_dataset}
\end{figure*}

Figure~\ref{fig:our_dataset} shows scenes in our dataset.
These data \frev{are available on the project website.}\footnote{\protect\url{http://www.computationalimaging.org/publications/nlr/}}  

\paragraph{DTU}
We use the multiview images and camera calibrations provided by \cite{jensen2014large} along with object masks by Niemeyer~\etal \cite{Niemeyer2020CVPR} and Yariv~\etal \cite{yariv2020multiview}.
Each scan consists of 49 or 64 images with 1600\,$\times$\,1200 pixels and covering approximately 90\,$\times$\,90 degree view zone of the object. 
Ground truth 3D shapes in form of point captured using structured light stereo are used to compute the Chamfer distance.

\paragraph{Volucap}
We evaluate a full body reconstruction using a sample frame from a video sequence provided by the volumetric production studio Volucap GmbH \cite{volucap2020}. 
The cameras are distributed in pairs around an upper hemisphere of the capture dome with a subject standing in the center. 
The images are cropped and rescaled to 2028\,$\times$\,1196 px. The masks are manually annotated on top of automatic background subtraction.

\paragraph{Digital Ira}
The Digital Ira dataset \cite{fyffe2014scu} contains 7 very high resolution (3456\,$\times$\,5184 px) closeups of a man's face in a studio setting. We used the provided camera calibration data and we manually annotated objects masks for each view.

\section{Baselines}

\paragraph{Colmap} 
We follow the same procedure and settings for Colmap~\cite{schoenberger2016mvs} as Yariv~\etal~\cite{yariv2020multiview} to reconstruct a 3D point cloud from the multi-view images, build a surface mesh using Poisson reconstruction (with trim = 7) and render the images using PyRender.
For the DTU dataset, we also apply object masks to remove background points before the surface reconstruction.
Note that this method failed to produce a surface mesh of the Digital Ira \cite{fyffe2014scu} with all variations of parameters that we tried.

\paragraph{NeRF}
We use the original code shared by the authors \cite{mildenhall2020nerf}. 
We modify the code to support general camera models. 
We execute the code using the provided high-quality configuration used for their paper results but we reduce the number of random samples to 1024 to fit to our GPU memory (11 GB on Nvidia Geforce RTX 2080Ti). 
Further, to support variable sensor sizes in our dataset, we upsample all images to the shape of the high-resolution cameras which retains the detailed information. 

For our dataset, we are forced by the memory requirements to reduce the training resolution by factor of 2 from 3000\,$\times$\,4000 pixels to 1500\,$\times$\,2000 pixels.
However, it is unlikely that this reduction affected the results since our experiment with even more aggressive four-fold downsizing (to 750\,$\times$\,1000 pixels) yielded a PSNR of $28.62$\,dB which is comparable to $28.58$\,dB of the two-fold downsizing we ended up using.

\paragraph{IDR}
We use the original code and configuration shared by the authors \cite{yariv2020multiview}. 
Note, that while IDR supports training without accurate camera calibrations, we train with the same calibrations used for all other methods and keep the poses fixed during the process.

\paragraph{Neural Volumes}
We use the original code shared by the authors \cite{Lombardi:2019}. 
We train our models for at least 150K iterations. 
As recommended by the authors, we use three views from different angles to condition the autoencoder. 
The authors provide the ability to either pass the background explicitly or to let the network estimate the background. Since we don't have a background image for all datasets, we choose to let the network estimate the background.
Due to the long training times, we evaluate the method on a limited subset of our dataset. 
We crop the lower resolution landscape images to the same aspect ratio as the six high-resolution central cameras in portrait mode.
Due to GPU memory limitations, we are only able to train with the image resolution reduced to $1/4$ of the original size of the high-resolution images. 
Note that Neural Volumes uses an explicit voxel representation and is therefore, unlike implicit models, not resolution independent.

\section{Metrics}

\paragraph{PSNR}
For all methods we compute the PSNR between the reconstructed images and the ground truth only for pixels in the object masks.
The same procedure is used by Yariv~\etal~\cite{yariv2020multiview} and we verify it by reproducing their metric scores with their pre-trained models.
Therefore, even methods that do not use masks to recover the shape (NeRF, Neural Volumes) are only rated based on prediction of the foreground object.

\paragraph{Chamfer distance}

The Chamfer distance is computed in a similar manner as in the original Matlab scripts provided with the DTU dataset \cite{jensen2014large}.
\frev{The shortest distance between each ground-truth 3D point and the reconstructed shape} is computed in one direction by the Open3D \cite{Zhou2018} library for Python.
This prevents penalizing reconstruction of background (NeRF) or reconstruction of occluded object parts.
We leave out Neural Volumes from evaluating the metric since surface extraction was not demonstrated in the original manuscript \cite{Lombardi:2019} and the explicit nature of the representation makes recovery of fine details difficult.

\section{Sphere tracing implementation details}

Convergence of the main paper Eq.~5 towards the zero-level set $\sdf_0$ is not guaranteed for general shape configuration.
As the SDF function $\sdf$ indicated distance towards the nearest surface, it underestimates the optimal step length if the nearest surface is not orthogonal to the current ray.
This can partially be improved by dividing the step length by the dot product $-\rd \cdot \nabla_{\coords} \sdf(\coords_i)$.
However, computing the gradients in every step is costly and quickly diminishes the returns.
Furthermore, if a ray closely misses a surface edge with a surface normal orthogonal to the ray direction, such adjustment does not improve the convergence as the step length will stay very small.

To mitigate these issues, we use bidirectional sphere tracing as described by Yariv~\etal \cite{yariv2020multiview}.
This approach does not rely on the sphere tracer's convergence as it uses it to only narrow a possible range of surface positions which is then further refined by additional solvers.

To this goal, we solve Eq.~5 to get the near zero-level set $\coords^n_n$ in a forward direction for $n = 16$. 
We mask rays with $|\sdf(\coords_n)|<5e^{-5}$ as converged. 
These rays are not further optimized and $\hat{\coords}_n = \coords^n_n$ is the final output of the \frev{sphere tracer}.
Next, for the remaining rays, we solve a modified equation in an opposite direction to find a far zero-level set $\coords^f_n$ as
\begin{align}
\coords^f_0 = \ro + t_{f} \rd , \quad \coords^f_{i+1} = \coords'_{i} - \sdf(\coords^f_i) \rd.
\end{align}
where $t_{f}$ is the rear intersection of $\rd$ with a unit sphere.
$\coords^n_n$ and $\coords^f_n$ define the nearest and the furthers possible location of the first surface point $\coords_n$.
In cases where $\coords^f_n$ is closer than $\coords^n_n$, we conclude that no surface exists.
In the remaining cases we evaluate 100 samples with the candidate range and look for the first zero crossing of the $\sdf$.
Finally, assuming a locally monotonic $\sdf$ we further refine the zero-crossing location by 8 steps of sectioning \cite{yariv2020multiview} to get the zero-level set $\hat{\coords}_n$.

Note that for performance, stability and memory efficiency, this procedure is not differentiated \cite{yariv2020multiview,jiang2020sdfdiff,liu2020dist}. 
Therefore, one extra step of the forward sphere tracing is executed with gradient tracking enabled to achieve a differentiable sphere tracing.
We apply the gradient direction adjustment \cite{yariv2020multiview} for this last step to get the final zero-level set as 
\begin{align}
\coords_n = \hat{\coords}_n - \frac{\sdf(\hat{\coords}_n)}{\nabla_\coords \sdf(\hat{\coords}_n) \cdot \rd} \rd,
\end{align}
for points where $|\sdf(\coords_n)|<0.005$.
A division by zero is avoided by clamping the denominator to $|\nabla_\coords \sdf(\hat{\coords}_n) \cdot \rd| > 0.01$.

\end{document}